\documentclass{article}
\usepackage[utf8]{inputenc}
\usepackage{comment}
\usepackage{graphicx}
\usepackage{geometry}
\usepackage{float}
\usepackage{array}
\usepackage{amsmath,amssymb,amstext}
\usepackage[parfill]{parskip}
\usepackage{lineno}
\usepackage{mathtools}
\usepackage{soul}
\usepackage{subcaption}
\usepackage{graphicx}
\usepackage[T1]{fontenc}
\usepackage[export]{adjustbox}
\usepackage[hidelinks]{hyperref}
\usepackage{cleveref}
\usepackage{tabularx}
\usepackage{indentfirst}
\title{Textured As-Is BIM via GIS-informed Point Cloud Segmentation}
\author{Mohamed Said Helmy Alabassy*}

\begin{document}
	\maketitle
	\begin{center}
		{*}\textbf{Affiliation:}~Bauhaus University Weimar, Faculty of Civil and Environmental Engineering, Chair of Construction Chemistry and Polymer Materials\\
		\textbf{Email:}~\href{mohamed.said.helmy.alabassy@uni-weimar.de}{mohamed.said.helmy.alabassy@uni-weimar.de}
	\end{center}
	
	This manuscript is an extended version of the following conference paper: Krischler, Judith; Alabassy, Mohamed Said Helmy; Koch, Christian (2023). BIM Integration for Automated Identification of Relevant Geo-Context Information via Point Cloud Segmentation. 30th International Workshop of the European Group for Intelligent Computing in Engineering (EG-ICE). Date: 04.07. - 07.07.2023, London, UK. \href{https://www.ucl.ac.uk/bartlett/construction/sites/bartlett_construction/files/bim_integration_for_automated_identification_of_relevant_geo-context.pdf}{URL}
	
	\setlength\parindent{15pt}
	\section*{Abstract}
	Creating as-is models from scratch is to this day still a time- and money-consuming task due to its high manual effort. Therefore, projects, especially those with a big spatial extent, could profit from automating the process of creating semantically rich 3D geometries from surveying data such as Point Cloud Data (PCD). An automation can be achieved by using Machine and Deep Learning Models for object recognition and semantic segmentation of PCD. As PCDs do not usually include more than the mere position and RGB colour values of points, tapping into semantically enriched Geoinformation System (GIS) data can be used to enhance the process of creating meaningful as-is models. This paper presents a methodology, an implementation framework and a proof of concept for the automated generation of GIS-informed and BIM-ready as-is Building Information Models (BIM) for railway projects. The results show a high potential for cost savings and reveal the unemployed resources of freely accessible GIS data within.
	
	\textbf{Keywords}~{Semantic Segmentation; Point Cloud Segmentation; As-Is Model; GIS-informed BIM; Scan2BIM; Railway Alignment; IFC}  
		
	\vspace{3pt}
	\textbf{Highlights}
	\begin{itemize}
		\item{Recreating horizontal and vertical alignments from semantically segmented railway trackbeds in GIS-compatible data format (i.e., GeoJSON) automatically.}
		\item{Developing a pipeline to generate UV-mapped coloured textured meshes of irregular 3D shapes automatically from vertex coloured point clouds with embedded texture atlas using IFC.}
		\item{Showcasing potentials for cost savings in early planning phases of railway projects by relying on free and open GIS data. Automating the creation of Alignments in IFC in combination with accessibility to geographic databases within BIM could provide further insights relevant to railway planning activties by querying classes or identifying multi- or bi-temporal differences in the vicinity of railway lines.}
		\item{Demonstrating the technical feasibility of progressive UV-mapped textured meshes compatible with the IFC data format and visualising their information correctly within currently available free and open source BIM software, which should reduce the loss of semantical information exchanged between CityGML 3 and IFC and benefit other BIM related applications, like information modelling of damages.}
	\end{itemize}

	\section{Introduction}
\label{intro}
\subsection{Motivation}
Due to the ailing state of the German railways, an increased focus on digitisation is directed recently to maintain, upgrade and expand the existing railway network and prepare it for the future~\cite{DBInfraGOAG.2024,Dickenbrok.2024}. Digitisation of the standing network is a very costly and time-consuming endeavour, especially in Germany, where over 33,000 km of railway exist, with almost 2,000 km of railway tracks, 200 bridges and 18,000 points and crossings renovated and upgraded in 2023 alone~\cite{DeutscheBahnAG.2023}. This is even more pronounced in the early stages of the larger construction projects, upon which status-quo data (e.g., plans, drawings, or surveys) depends, that is often incomplete, exists in non-digital formats or could be missing altogether~\cite{DBAG.2022}. However, this data still needs to be processed for integration into inter-compatible working methods based on Building Information Modelling (BIM)~\cite{Renzel.2021,Hofstadler.2021,DBEnC.2024}.

The time-frame of such processes from a project's initiation to the finishing handover often spans over decades as evident from some of the German Reunification Railway Transport Projects number 8 and 9 for the expansion and new construction of the routes Nuremberg\textendash Erfurt\textendash Halle/Leipzig\textendash Berlin, and along the route between Leipzig \textendash Dresden respectively~\cite{ScahstandVDE.2023}, where the acquisition of up-to-date surveying data of the location is crucial to ensure correct planning and informed decision-making. Especially through the very inceptive phase of the project, when there is often little to no allocated funds to support the decisions needed to be made that are in turn time-varying on shifting requirements, demand, possible project variants, extents, costs, regulations and permits, etc~\cite{Haghsheno.2024}.

A patchy data basis, however incomplete it might be, could be improved upon with registered surveying patches for regions of interest using commercial aerial Light Detection and Ranging (LiDAR) scanning services, through photorealistic 3D tiles from the latest Google Maps 3D immersive view~\cite{googlemaps.2024} or other similar alternatives. Yet, surveying services are often very expensive, rendering their costs a hindrance to utilise, as they are neither always available, easy to acquire at remote locations nor guaranteed to be up to date over time with changes of the status-quo on ground. Thus, a more practical and less costly approach, that includes openly accessible geographic and location information free of charge provided by online web mapping platforms and Geographic Information Systems (GIS), could be of benefit. The open geodata published by official surveying authorities and available for free could be an economical alternative source to acquire a relatively up-to-date data basis that can be processed and utilised to generate as-is BIM models.

The latest strides in the research field of intelligent recognition of Point Cloud Data (PCD) have been established in state of the art systems for autonomous driving, navigation, flight planning for drones, and aerial scans' segmentation, that could be utilised for the semantic segmentation of point clouds generated from available geodata. This would help classify further objects of interest that are unlabelled in the raw data or difficult to attain otherwise to derive further insightful knowledge and semantic information that assist in the identification of interdependencies and relations between labelled classes and project requirements defined by official regulations, public authorities, planners and other concerned stakeholders.

Furthermore, the various types of data and information formats, compounded by limitations to incompatible software and hardware used by each actor, emphasises the benefits to common adoption of freely and widely available open data formats, which are information-lossless, convertible and inter-compatible with technical hardware and software resources of all involved stakeholders. This early phase always requires extensive data exchange between various stakeholders, that highlights the importance of adopting big open BIM to facilitate interoperability and reduce the incompatibility of different data formats separately used by each party involved.

Such reconstruction of the status-quo, as relatively accurate BIM-ready as-is railway models they are, using freely available PCD (i.e., Scan2BIM) suffices to propel information based decision-making forward in early planning phases immensely with minimum cost. A hybrid approach that incorporates not only PCD, but also geospatial and a mix of other free publicly accessible data may help overcome common obstacles arising from insufficient data basis.
\subsection{Research Question}
Hence, this study considers the applicability of state of the art methods available today to find practical solutions to the main question of how to optimally combine the wealth of information available in open geodata, including the German official topographic and property cadastre information systems (i.e., photos, point clouds, property and land use shapes and attributes) through semantic segmentation of point clouds. The focus lies on the railway infrastructure domain, with BIM models in the open IFC format, which paves the way for a very economical workflow to reconstruct an as-is information model well suited for BIM-based planning in early project phases. It focuses on the transfer of knowledge attained from semantic PCD segmentation through surface reconstructed geometry and derived semantics and properties into big open BIM. We hypothesise that the inclusion of GIS data may help overcome technical and financial limitations to insufficient data basis. This research further expands on previous work by the authors~\cite{Krischler.2023}. For this purpose, we rely solely on freely available data, using only open-source tools and convert our results into the open BIM IFC data format.
\subsection{Main Contributions}
A modified version of the deep neural network (DNN) architecture 2DPASS has been used to train a model capable of semantically segmenting buildings, vegetation, water bodies, roads, and railway trackbeds. The used architecture relies in the training phase both on 3D features from a dataset of processed and semi-automatically annotated LiDAR scans as well as 2D features from coloured orthophotos. The original dataset is acquired from freely available LiDAR scans, digital orthophotos, ATKIS shape files from the geoportal websites of the Free States of Thuringia and Saxony. The authors of this study attempt to make the most use of free and open data sources, software, and information formats for pre- and post-processing of point cloud data, semantic segmentation, post processing of the inferred segmentation results with 3D reconstruction and modelling of railway related contexts.

The contribution of this work links to formerly published literature about Scan2BIM approaches in railway~\cite{Soilan.2020,Ariyachandra.2021,Eickeler.2021,Cserep.2022,Ariyachandra.2023}. The proof of concept to our proposals is demonstrated on two case studies. The first case study demonstrates the potentials for cost savings in early planning phases of railway projects by estimating horizontal and vertical alignments from the inferred semantic segmentation of the predicted railway trackbeds in GIS-compatible data format (i.e., GeoJSON) that could be later used as input for automatically generating and \textit{IfcAlignment} or querying multi- or bi-temporal differences along a buffer zone of the alignment.

The second case study demonstrates the texturing potentials of the resulting IFC files and builds on former studies by \cite{Huthwohl.2018,Sacks.2018,Artus.2022}, and expands it further as no attempt has been found in the body of literature so far to generate UV-mapped textured meshes of irregular 3D shapes in IFC automatically, whereas the aforementioned studies successfully embedded a texture image onto a 2D flat surface in 3D space with manual UV-mapping.  
	\section{Related Works}
\label{related_works}

\subsection{Point Cloud Segmentation}
Point cloud acquisition has become more ubiquitous due to advancements in latest depth and light capturing sensors, that could be mass-produced within a wide range of price, quality and application spanning from LiDAR scans in engineering surveying to those fitted on drones or the miniaturised variants in smartphones and extended reality (XR) gadgets. This wide adoption has driven progress on semantic segmentation of point clouds to develop highly sophisticated Deep Learning (DL) models able to classify large amounts of points automatically on graphical and computer processing units for a wide range of applications ranging from early phase planning of projects, scene understanding for autonomous driving to construction monitoring and inspection, XR applications and beyond.

Texture utilising methods for point cloud segmentation involve a collection of varying approaches. One focuses on fusing representations from points, voxels, and/or projection images within different branches of the network’s architecture. Tang et al. combined point-wise MLPs in each sparse convolution block to learn a point-voxel representation~\cite{Tang.2020} and relied on Neural Architecture Search to reach an optimally efficient design . Xu et al. proposed a range-point-voxel fusion network (i.e., RPVNet) to utilise information from the three representations~\cite{Xu.2021}. However, texture features and semantics from photos were underutilised or discarded altogether by focusing only on colourless LiDAR point clouds.

Another approach fuses inputs from multiple sensors leveraging benefits of both camera and LiDAR~\cite{Madawy.2019, Krispel.2020, Meyer.2019}. El Madawy et al. converted RGB images to a polar-grid mapping representation and designed early and mid-level fusion architectures~\cite{Madawy.2019}. Point Painting exploited the segmentation logits of images~\cite{Vora.2019} and projected them into LiDAR space by spherical~\cite{Milioto.11320191182019} or bird’s-eye projection~\cite{Yuan.2018} for LiDAR network performance improvement, whereas Zhuang et al. exploited a collaborative fusion of two modalities in camera coordinates~\cite{Zhuang.2021}. Yet, the paired multi-modality data requires multi-sensor inputs in both training and inference and is more computationally intensive and unavailable in practical applications.

A third approach involves knowledge distillation, which is a technique used to transfer knowledge learnt by compressing a large teacher network to a smaller student~\cite{Hinton.2015}. Further improvements to knowledge transferring were achieved using different methods of matching feature representations~\cite{Ba.2013,Chen.2017}. For instance, aligning attention maps~\cite{Zagoruyko.2016} and Jacobean matrices~\cite{Srinivas.2018} were independently applied. This technique has been applied successfully to transfer priors across different modalities by using additional images in the training phase and to improve performance at inference~\cite{Gupta.2015,Wang.2018,Yuan.2018b,Liu.2021,Zhao.2020}. For instance, by either inflating kernels of 2D convolution into 3D~\cite{Xu.2021}, introducing a 2D pixel-to-point assisted pre-training~\cite{Liu.2021b} or by utilising a teacher-student framework~\cite{Yuan.2022}. Furthermore, 2DPASS leveraged auxiliary modal fusion to transfer 2D knowledge through multi-scale fusion-to-single knowledge distillation, which took care of the modal-specific knowledge and demonstrated generality, flexibility and effectiveness when compared to other fusion based methods~\cite{Yan.2022}.
\subsection{Scan2BIM in Railway Infrastructure}
The so-called Scan2BIM approach means the (automated) acquisition of PCD and the subsequent creation of BIM models from it ~\cite{Borrmann.2021}. Many works deal with the topic of applying Scan2BIM to railway infrastructure. This subsection presents an overview of the related research.

Ariyachandra et al.~\cite{Ariyachandra.2021} described a method for creating large-scale geometric information models (GIM) of rail infrastructure using automated segmentation of point cloud data (PCD). The geometry of the segmented rails and track bed was then reconstructed resulting in IFC files of the respective objects. The authors distinguished between a distinct segmentation of rails and track beds. In their earlier publications~\cite{Ariyachandra.2020}, they focused on the identification of railway masts of double-track railways from LiDAR PCD by taking railway design rules strongly into account and therefore renouncing neighbourhood structures, scanning geometry and the intensity of input data. After removing the adjacent vegetation, a track corridor remains from which the masts were then extracted, which are parallel to the global Z-axis. The masts were then differentiated from other pole-like objects defining inner and outer boxes around the masts to identify outliers. The proposed method reached high detection and precision rates, and included eventually the 3D mast models in IFC.

Yang~\cite{Yang.2014} extracted rail beds and rails from mobile laser scanning (MLS) point cloud. They were able to detect different rail segments (where one alignment ends and the other begins, e.g., points) and reached an overall detection accuracy of more than 95\% in completeness and quality and 99.78\% correctness. The applied method used cross-sections and so called ``scanning lines'' to identify the ballast bed of the rail and with the help of the specific geometric angle of the ballast bed, and was able to identify the course of the railway and to identify objects that did not belong to the railway itself. To avoid scanning the whole point cloud, the areas of interest have been segmented in the point cloud. In order to quantify the accuracy of the result, rail lines were manually digitised as lines and with a buffer zone of 15 cm compared to the extracted railway lines from the point cloud.

Cheng et al.~\cite{Cheng.2019} used coloured point clouds (range accuracy $\pm 2~mm$) in order to first segment them into (single-track) railway bed and remaining points. The tunnel cross-sections were extracted and the parameters derived. Ten different classes were identified, including rail head and railway bed. To estimate the horizontal and vertical alignments of the railway, a method of 3D local cylindrical neighbourhood was used. The presented method for classifying point cloud reached close to 100 \% accuracy. The authors achieved to create a parametric BIM model from the derived objects, yet it was not converted into an open product data model.

Eickeler and Borrmann~\cite{Eickeler.2021} presented a three-stage concept for the creation of a railway specific dataset that allows a higher grade of automation in railway asset detection. The paper focused on videos and images, not on PCD derived from LiDAR scans. The three stages consisted of 1. Simple Class Annotation (SCA), 2. classification by domain knowledge and 3. creation a full asset model (optimisation).

Cser{\'e}p et al.~\cite{Cserep.2022} used coloured LiDAR PCD from a vehicle-mounted scanner (mobile mapping system; 60 km/h). The PCD was fragmented and then used for cable and rail recognition. To remove vegetation within the PCD along the railway tracks, contour detection was used. The railway was identified using contour finding and Hough transformation. The paper compared three different algorithms for detection of the railways. First, the trackbed was detected taking into account the point density and the most common height in a defined area. Second, the PCD was reduced to a 2D digital elevation model (image) and the 2D Hough transformation used in order to identify rail pair from the resulting 2D projected PCD. With this methodology of creating as-is models, it was hard to detect cables being placed underneath each other, but the standardised track gauge could be taken into account. The 3D line of the railway was detected using Hough transformation. Third, the region growing algorithm with the trajectory of the identified cables was used. The trajectory of the power cables was calculated with the Random Sample Consensus (RANSAC) algorithm. All three approaches have been compared. For the cable objects, the Region growing algorithm showed the best accuracy and operated the fastest.

The work of Soil{\'a}n et al.~\cite{Soilan.2021} included an automated approach to derive precise railway alignment (centreline between two tracks) from LiDAR scans in order to create IFC alignment objects from that. The authors used LAS files (without colour) created by ground vehicle mounted LiDAR sensors (2000 points per square metre) to extract railway alignments and expressed them as IFC files (xBIM toolkit for IFC 4x1 was used). The point cloud was segmented in the direction of the trajectory using a distance filter in order to minimise the processing effort. An intensity-height-filter was applied to remove non-rail points. The extracted railway line was stored as polylines and the centreline between the rails was used to create the \textit{IfcAlignment} instance to form an IFC file. The validation was carried out comparing the generated alignment using the reference points for the construction.

The reconstructing of 3D railway geometries from surveying data often depends on highly detailed point clouds that can even serve as a basis for the extraction of overhead cables. All of the presented works required highly detailed point clouds and have often created the data sets for the sole purpose of reconstructing highly detailed geometries. Some of the reviewed work has eventually converted the 3D reconstructions to the open IFC data model. None of the found work used GIS data to support the semantic segmentation process.
\subsection{Texturing and Semantic Enrichment}
Hüthwohl et al. successfully demonstrated the applicability of mapping a 2D texture to a 2D plane for a beam face of a 3D IFC model~\cite{Huthwohl.2018} using an image as a 2D texture in 3D space, albeit in the context of classification of Building Information Modelling (BIM) of damages and exploring their suitability to include inspection and integrating damage information modelling for RC bridges in a standardised and open IFC format within an open bridge management system~\cite{Sacks.2016, Sacks.2018}. A prototype software was developed for that purpose, allowed by the use of ``Gygax construction IT research platform for 2D \& 3D''~\cite{huthwohl2017gygax}, that relied on IFCEngine DLL, Helix Toolkit and SharpDX, to allow the shader to render the texture as a UV-mapped (i.e., S,T in IFC) triangulated mesh consisting of split rectangle into two triangles to a linked image.

The authors highlighted several disadvantages to this approach due to problems with estimating the correct location and orientation of the image plane in the IFC model, the distortions resulting from taking close-up photos and increase in data size by embedding the texture into the IFC model. This furthermore includes the risk of corrupting the IFC file during data exchange, when an external Uniform Resource Identifier (URI) of the image embedded in the IFC model without sharing the attached texture or when the texture image is moved to a different location.

Several pilot attempts for formalising compatible texture generation and integrating it within the entities of the IFC schema have been discussed in grey literature and shown potentials for BlenderBIM's capability to handle texture maps~\cite{Moult.2022}. However, no attempt has been found in the body of literature so far to generate UV-mapped textured meshes of irregular 3D shapes in IFC automatically, mainly due to the complex set of challenges to be faced from aliasing problems arising from inadequate sampling of all colour frequency values for texels through interpolation of texture attributes projected onto an intermediary arbitrary parameterised surface. Mipmapping based on Level of Development (LoD) requirements could be a solution in this case, yet it would require generating at least 3 texture maps of different resolutions based on the required triangle area of the triangulated faces defined for each LoD 3 and upwards.
\subsection{Research Objectives}
\begin{itemize}
\item{Examining the potentials for cost savings by using freely available GIS data in early planning phases of railway projects by estimating horizontal and vertical alignments from the inferred semantic segmentation of the predicted railway trackbeds in GIS-compatible data format (i.e., GeoJSON) that could be later used as input for automatically generating and IfcAlignment or querying multi- or bi-temporal differences along a buffer zone of the alignment.}
\item{Examining the feasibility and compatibility of progressive meshing and surface mapping from vertex coloured point cloud to generate UV-mapped textured meshes of complex irregular 3D shapes in IFC automatically.}
\end{itemize}

	\section{Methods}
\label{}
This section describes a methodology of using GIS-informed point cloud segmentation for an automated attempt of creating as-is models for early planning phases. Openly accessible Point Cloud Data (i.e., LiDAR), coloured orthophotos and 2D GIS data serve as a basis of the dataset for training the point cloud segmentation model. 
\begin{figure}[H]
	\centering
	\includegraphics[width=0.6\textwidth]{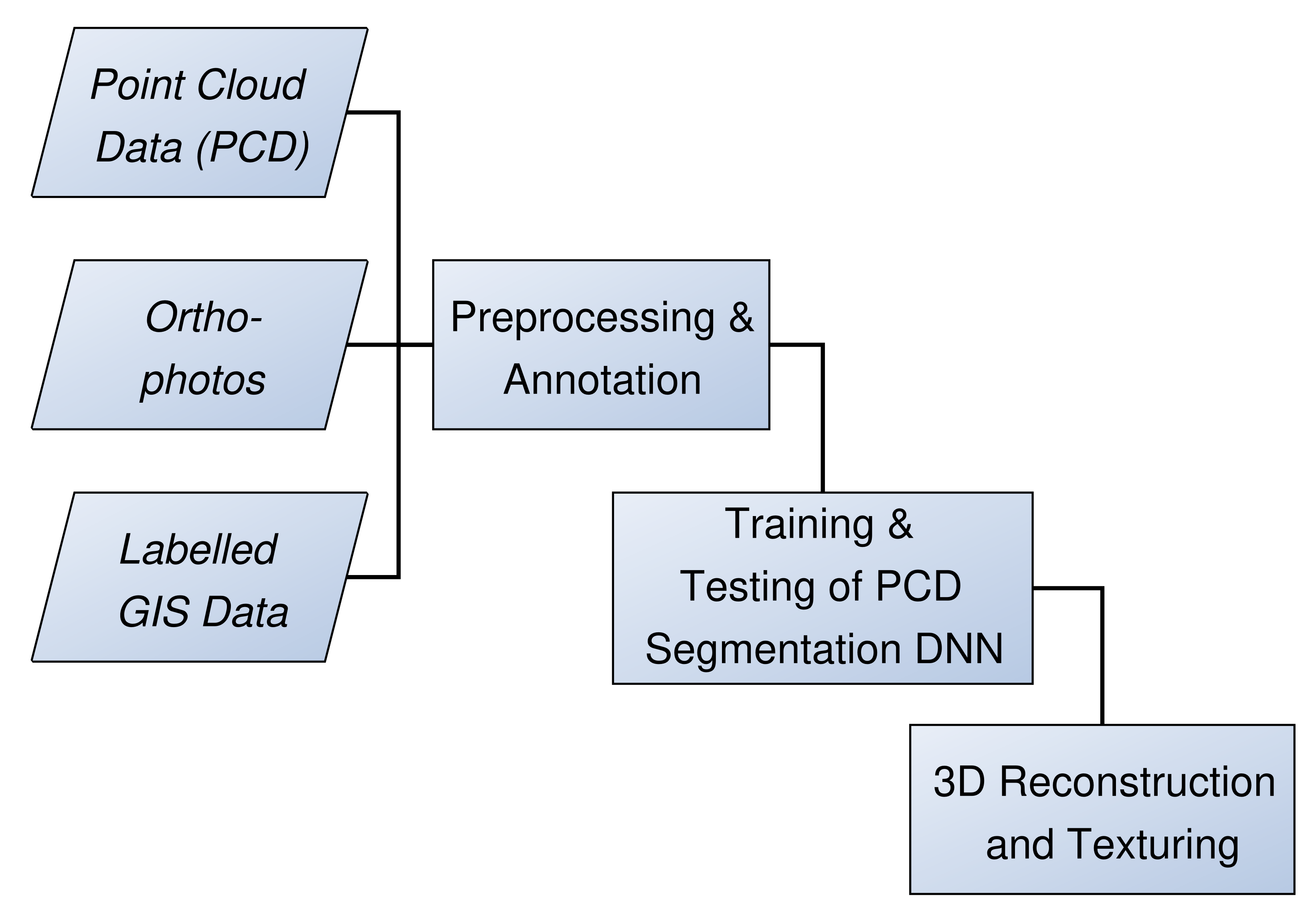}
	\caption{Simplified methodology of textured as-is modelling from GIS-informed semantic segmentation of PCDs.}
	\label{fig:methodology}
\end{figure}
\subsection{Input Data}
Usually, the LiDAR PCD consists of the x, y, z coordinates of points, CRS Projection Information and little to no semantic classification to them. Often, there are also no colours like RGB-values assigned to the points. Furthermore, PCD is often documented within tabular or textural formats, such as .csv, .xyz or within the widely used open format LAS, which may include some semantic classification, depending on the age, technical specifications of the  scanning device and its support for the latest LAS format version. The targeted PCD converted from a LiDAR scan in this study usually comes from surveying flights and is therefore, unlike data from other kinds of acquisition, recorded from above and therefore involving specific limitations that need to be addressed, which will be discussed in~\cref{implementation}.
\subsection{Preprocessing and Semi-automated Annotation}
In the first step, the orthophotos are used to colour the PCD, in the case of missing RGB values of the respective points. Orthophotos result usually from surveying flights and are typically provided as georeferenced, grid-based image data, such as .(geo)tiff.

The GIS data, together with its classification of GIS features, can be derived from querying GIS databases, such as OpenStreetMap or provided GIS data from public institutions. It is important to identify the necessary label and later object classes in advance to decide on which object classes should be represented in the subsequent as-is model. The used GIS data consists usually of 2D polygons, sometimes containing attributed information about the height of the real object, for example in the case of buildings. 
The labelling data that comes with the 2D GIS features, such as the classification into ``road'', ``vegetation area'', ``railways'', etc., is used in a next step to create annotation masks. Especially when using PCD originating from aerial scans, points with a higher z-value can cover other points of a different class, such as treetops covering a street when viewed from above. The GIS data helps to form coherently connected segmentation areas while also including algorithms of nearest neighbour to avoid false labelling. More information regarding this can be found in~\cref{implementation}. 
\subsection{Point Cloud Segmentation}
The annotation masks can now be used in a 80:20 training vs. validation dataset relation within the Deep Learning (DL) model, namely 2DPASS~\cite{Yan.2022}. The trained model can then be applied via inference to selected case study data, which is formerly unknown to the model. The outcome of applying the trained DL model to real data is an inferred segmentation from the already trained model based on a thresholded probability map. The resulting segmentation can be segregated into preassigned classes for reducing processing time and resources and their respective point clouds can then be treated individually.
\subsection{3D Reconstruction and Texturing}
Using the results from the previous step, the post-processing steps can be executed. This includes 3D mesh reconstruction from the class-specific point clouds with semantic information coming from the initial GIS data. Depending on the class, the result leads to 3D meshes of buildings in LoD2 or higher depending on the required Level of Information Need (LoIN) within a given coordinate reference system. The final meshes are furthermore first instanced as individual objects and then converted to a suitable data model, such as IFC and textured according to the orthographic photos.
	\section{Implementation}
\label{implementation}
\begin{figure}[H]
	\centering
	\includegraphics[width=\linewidth]{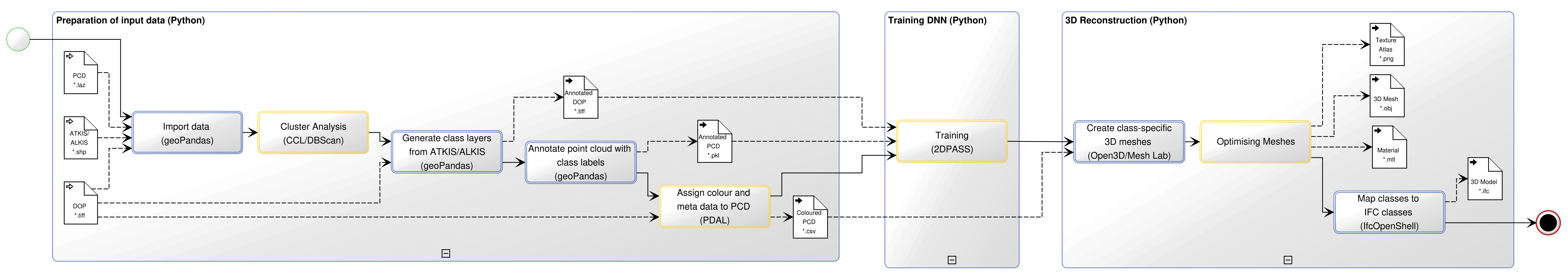}
	\caption{Detailed process map elaborating the methods used.}
	\label{fig:process_map}
\end{figure}

The developed pipeline for this study was mainly implemented in Python and minor portions through Python bindings to packages in C++ or background processes. \Cref{fig:process_map} showcases the steps of the workflow undertaken starting with downloading the original data directly from the geoportals of the States of Thuringia and Saxony that consists of the coloured orthophotos, LiDAR scans, cadastre masks, CityGML models of the cities of Erfurt, Jena and Weimar in the State of Thuringia, as well as Dresden and Leipzig in the State of Saxony respectively and annotating them semi-automatically then proof-checked manually. \Cref{tab:table_1} elaborates the various data types and formates included in the acquired dataset.
\begin{table}[H]
	\centering
	\caption{Freely available input data used within the implementation and case-study.}
	\label{tab:table_1}
	\newcolumntype{c}{>{\centering\arraybackslash}X}
	\begin{tabularx}{\textwidth}{ccc}
		\hline
		\textbf{Data}						& \textbf{Format}	& \textbf{Description}		\\
		\hline
		Digital Surface Model (DSM)			&   *.laz			&  		3D PCD, uncoloured  \\
		DigitalElevation Model (DEM)		&   *.laz			&  		3D PCD, uncoloured  \\
		Digital Orthophotos (DOP)			&   *.tiff			&  		2D images, coloured \\
		Cadastral Maps (ALKIS)				&   *.shp			&  		2D vector data      \\
		Topographical Maps (ATKIS)			&   *.shp			&  		2D vector data      \\
		Buildings, LoD2 (CityGML1)			&   *.gml			&  		3D city model data  \\
		\hline
	\end{tabularx}
\end{table}

\subsection{Preprocessing and Annotation of Input Data}
The base dataset from LiDAR scans provided in the LAS data format. It consists mainly of three layers: ground, above ground and outliers (i.e., Ground, 20 and Unclassified in Thuringian LAS files). \Cref{tab:lass_classes} details the default classes available in LAS format. As relying solely on the cadastre masks without further preprocessing steps was found insufficient to correctly annotate points related to specific classes like points related to external walls and corners of buildings, roads and railways that may lie outside the perimeter of the shape instances, which cannot catch their fine details. Therefore, to automate the annotation process of the other classes, each tile is processed with Connected Components Extraction and Density-based spatial clustering of applications with noise (DBSCAN) to cluster the point cloud and rely on the initial clustering for set operations with overlaid masks from the ATKIS relevant shapes that are geometrically manipulated to create buffer layers for annotation with the overlapping label extracted from their attributes' table.

\begin{table}[H]
	\centering
	\caption{Values and meanings of default LAS classes in LiDAR scans of dataset.}
	\label{tab:lass_classes}
	\newcolumntype{c}{>{\centering\arraybackslash}X}
	\begin{tabularx}{\linewidth}{p{0.5cm} c p{0.5cm} c}
		\hline
		\textbf{LAS Class}	& \textbf{Meaning}	& \textbf{LAS Class} & \textbf{Meaning} \\
		\hline
		0                    & Created, Never Classified  &		10   				& Rail           \\			
		1                    & Unclassified               &		11   				& Road Surface   \\			
		2                    & Ground                     &		12   				& Reserved       \\			
		3                    & Low Vegetation             &		13   				& Wire Guard (Shield)  \\			
		4                    & Medium Vegetation          &		14   				& Wire Conductor (Phase)  \\			
		5                    & High Vegetation            &		15   				& Transmission Tower \\			
		6                    & Building                   &		16   				& Wire Structure Connector (Insulator) \\			
		7                    & Low Point (Noise)          &		17   				& Bridge Deck  \\			
		8                    & Reserved                   &		18   				& High Noise   \\			
		9                    & Water                      &		\textgreater{}18    & User Defined \\	
		\hline
	\end{tabularx}
\end{table}

\begin{figure}[h!]
	\centering
	\begin{subfigure}{0.4\linewidth}
		\includegraphics[width=\linewidth]{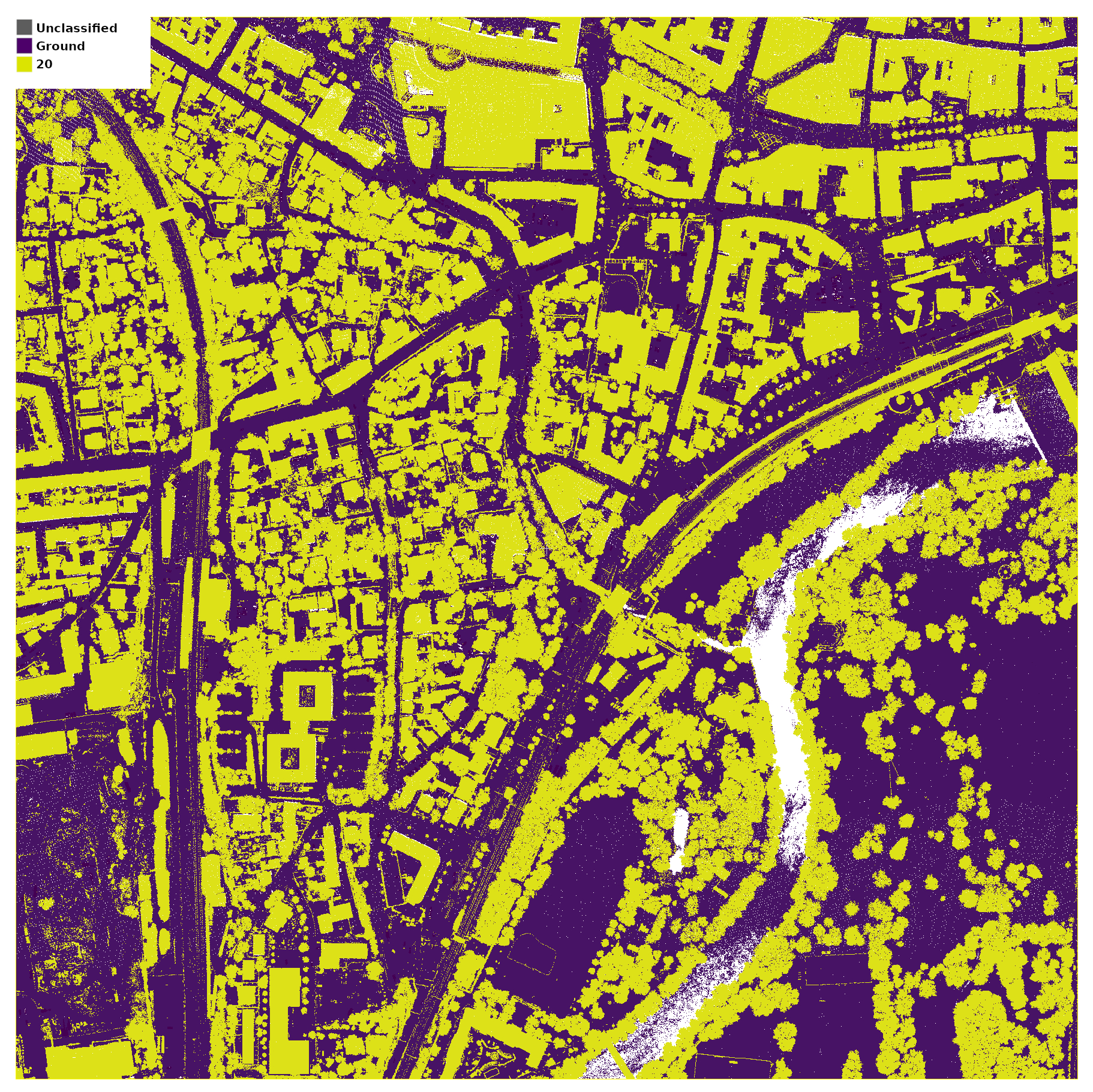}
		\caption{}
	\end{subfigure}
	\begin{subfigure}{0.4\linewidth}
		\includegraphics[width=\linewidth]{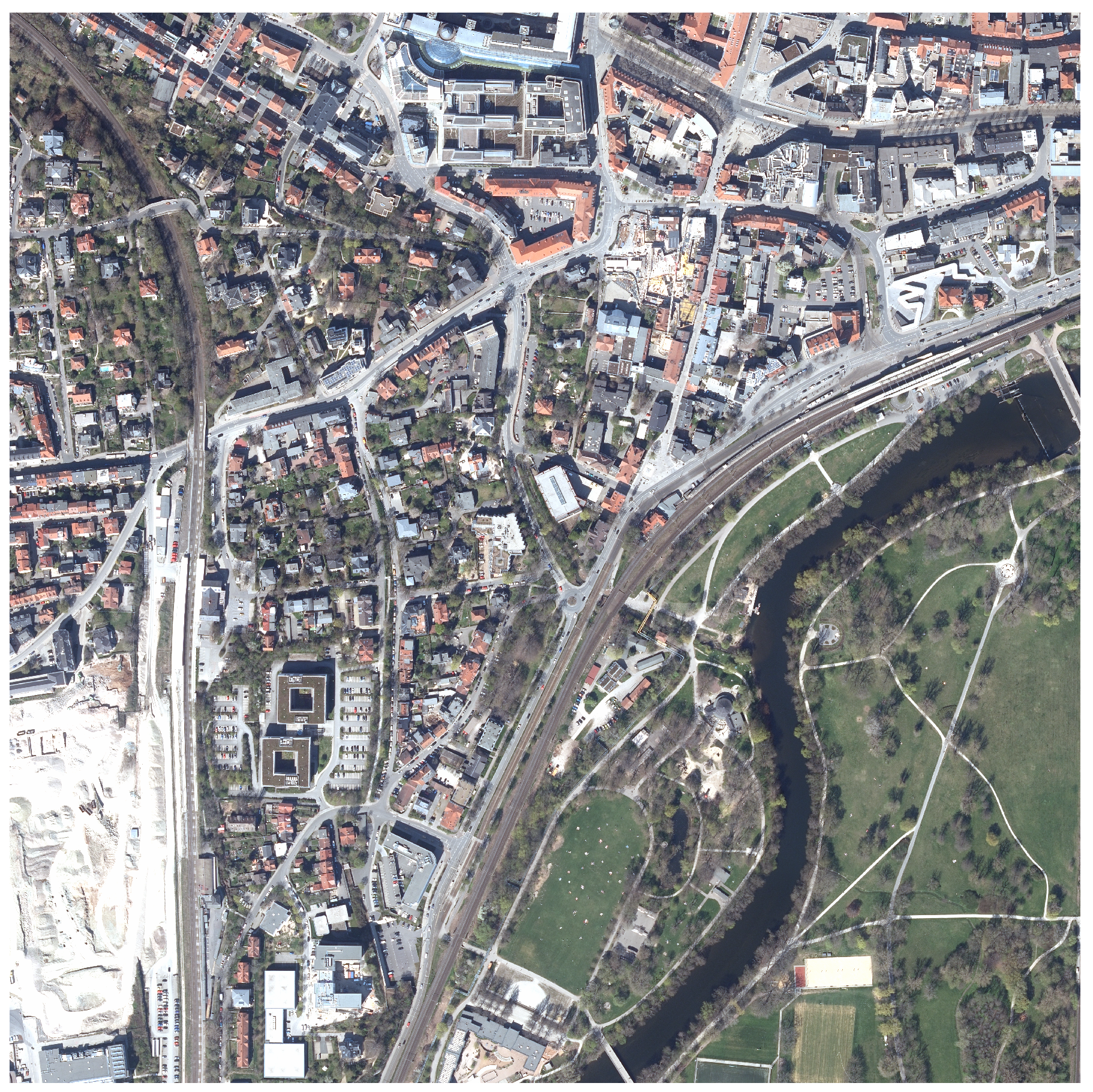}
		\caption{}
	\end{subfigure}
	\begin{subfigure}{0.8\linewidth}
		\includegraphics[width=\linewidth]{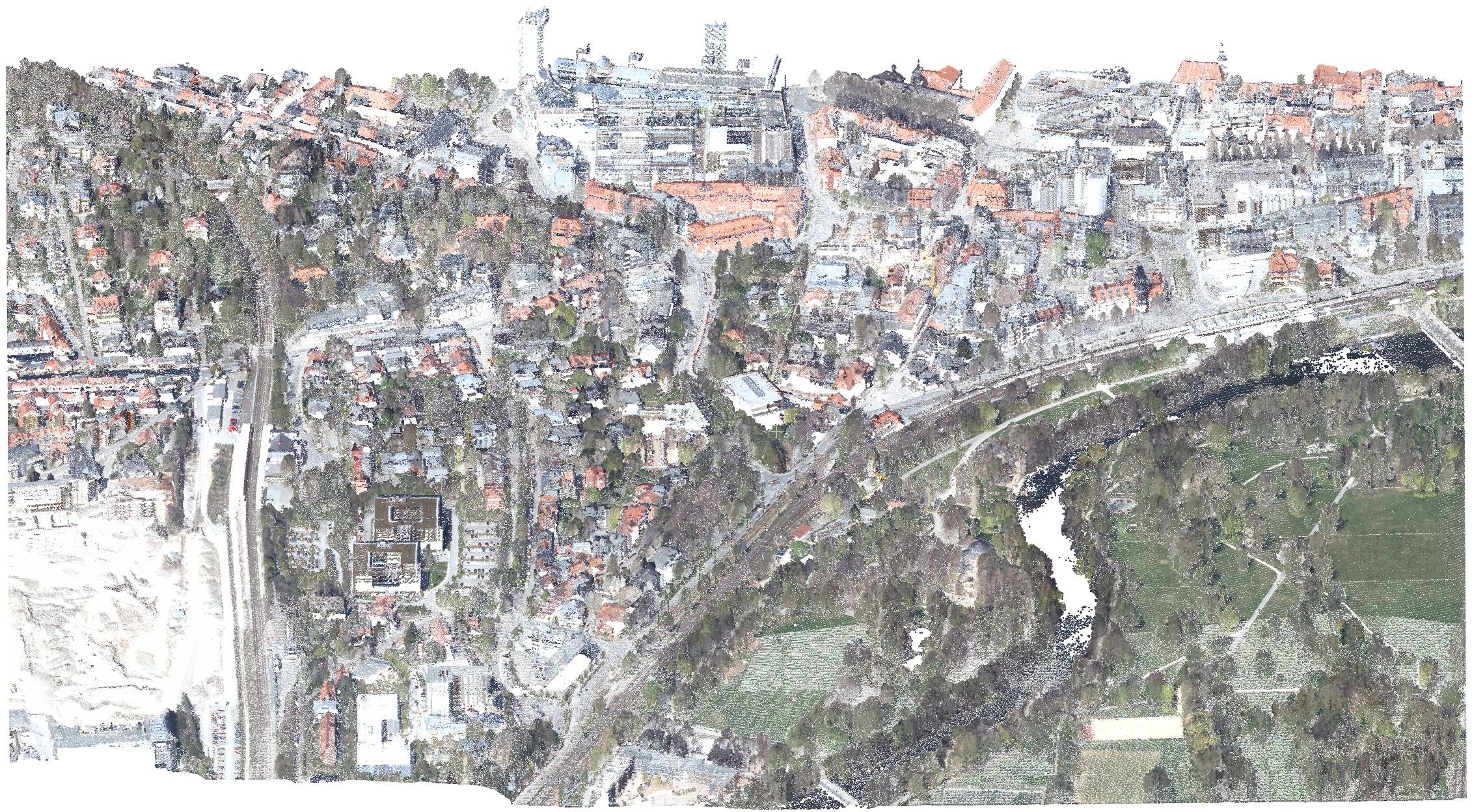}
		\caption{}
	\end{subfigure}
	\begin{subfigure}{0.266\linewidth}
		\includegraphics[width=\linewidth]{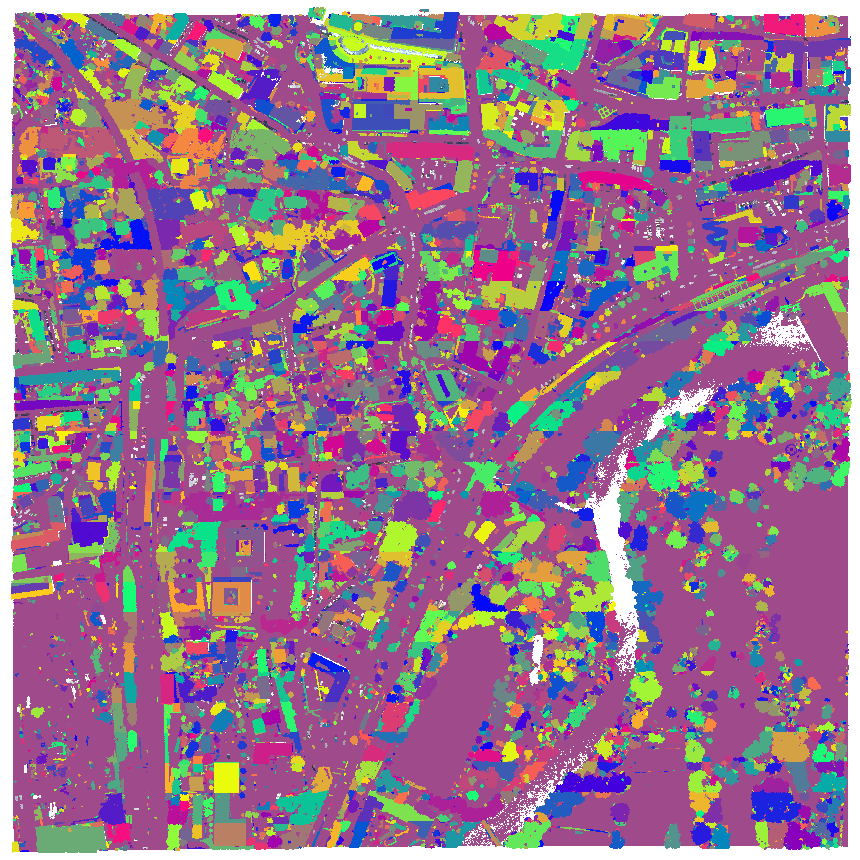} 
		\caption{}
	\end{subfigure}
	\begin{subfigure}{0.266\linewidth}
		\includegraphics[width=\linewidth]{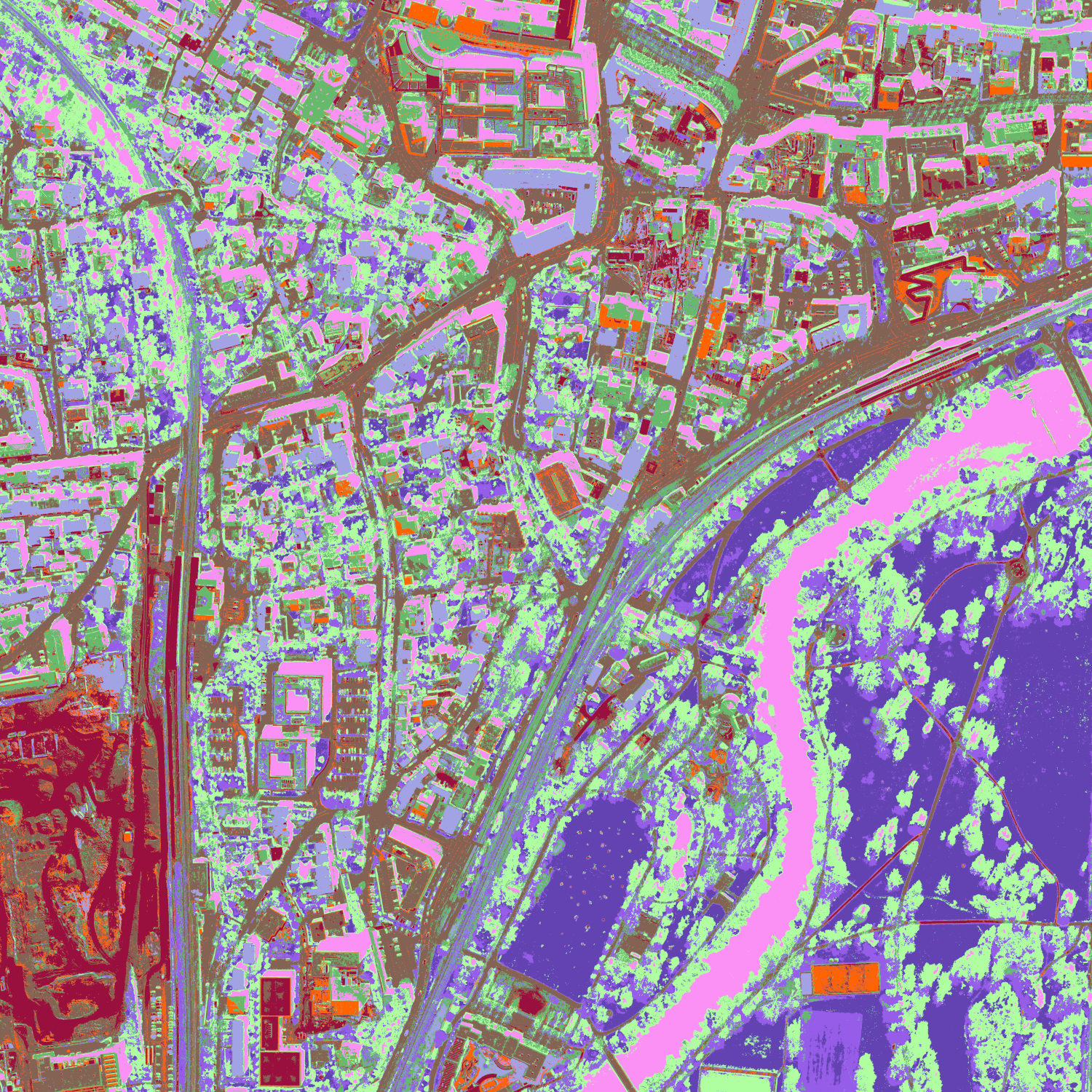} 
		\caption{}
	\end{subfigure}
	\begin{subfigure}{0.266\linewidth}
		\includegraphics[width=\linewidth]{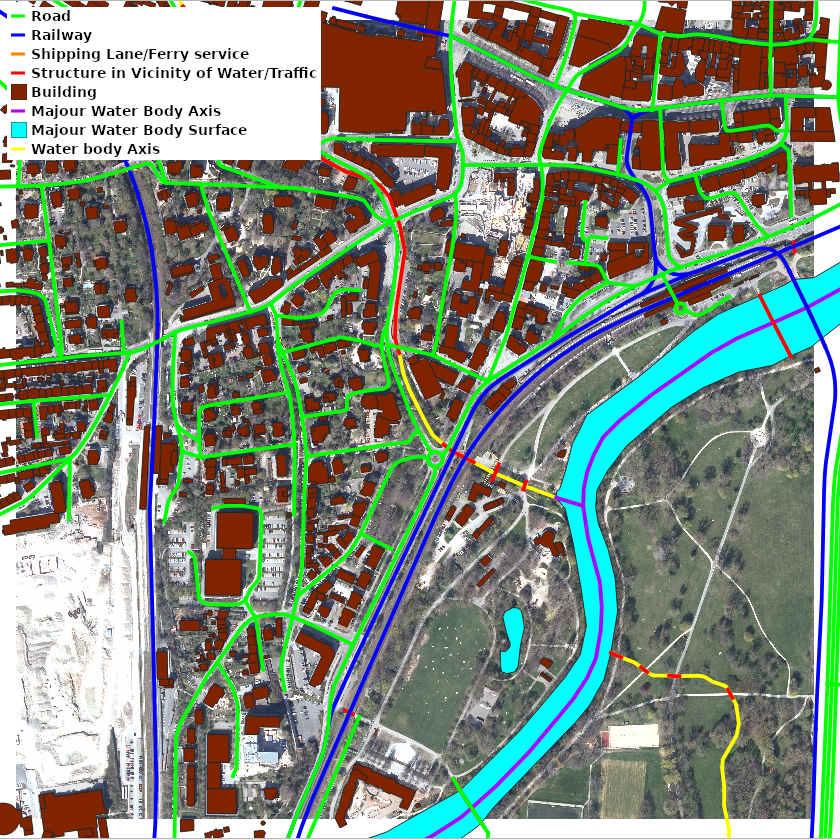} 
		\caption{}
	\end{subfigure}
	\caption{Overview of the original dataset tiles of LiDAR scans, Orthophotos and ATKIS mask and their initial pre-processing steps; (a) original LiDAR scan of a tile in Jena, Thuringia, (b) The tile's relevant orthophoto, (c) Colourised LAS file based on the sampling of colour from the orthophoto onto the vertices of the point cloud, (d) Connected Component Clustering of the LAS classes, (e) Labelled Gaussian mixture model fit to the orthophoto's pixels as features in RGB colour space, (f) ATKIS shape masks overlaid onto the orthophoto.}
	\label{fig:Dataset sample}
\end{figure}

The connected components labelling with CloudCompare operates by propagating a front on the surface implicitly represented by the point cloud with the Fast Marching algorithm and making it dependent on scalar distance values associated to each point and neighbouring entities within an octree structure; which is followed by smoothing via Gaussian distance field gradient~\cite{GirardeauMontaut.2006}. A sample of the dataset original PCD and orthophotos and the underlying processed clustering interim results for the semi-automated annotation are showcased in \cref{fig:Dataset sample}.

The cadastral shape files follow a very consistent and detailed data schema that is managed by the Working Committee of the Surveying Authorities of the States of the Federal Republic of Germany (AdV) and disseminated through the Documentation on the Modelling of Geoinformation of Official Surveying and Mapping (GeoInfoDoc; German: Dokumentation zur Modellierung der Geoinformationen des amtlichen Vermessungswesens (GeoInfoDok))~\cite{Portele.2022}. The adopted and currently used version 7.1.2 was used to query the relevant attributes from the transport mode attributes for rail and road and thus facilitate the automatic annotation of the data set.

The data structure is designed to store and organise GIS data within the so-called AAA schema to facilitate managing and querying spatial data~\cite{Portele.2022}. AAA stands for AFIS/ ALKIS/ATKIS; which refers to the official reference-point network, cadastre data and topographical maps. The geographical information contains the actual spatial data including geometric representations of geographic features such as points, lines, and polygons, which are stored in the Shapefile format (i.e., .shp). The associated attributes or properties of the geographic features are stored as key-value pairs with information about the keys and the values’ units derived from the GeoInfoDoc. The geodetic Coordinate Reference System (CRS) is ETRS89 for UTM zones 32N and 33N in the States of Thuringia and Saxony (i.e., EPSG 25832 and 25833) respectively. Information about the coordinate reference system used for the geometries, including its name, authority, and parameters. Since the number of attributes queried was very limited, there was no direct attempt to use prior knowledge about indexing the data to optimise for efficient spatial queries. For roadway axis object ``Ax{\textunderscore}Fahrbahnachse'' the following attributes were queried as listed in \cref{tab:roadway_attributes_table}.

\begin{table}[h!]
	\centering 
	\caption{Relevant roadway attributes extracted from the shape of the roadway axis ``Ax{\textunderscore}Fahrbahnachse'' object for labelling PCD. The first column constitutes the attributes' label, which is a three letter abbreviation of the attribute name in German. The second column provides the translated meaning into English and the third shows the data type of the values as either strings, integers or string enumerators (i.e., str, int, str enum) respectively.}
	\label{tab:roadway_attributes_table}
	\newcolumntype{c}{>{\centering\arraybackslash}X}
	\begin{tabularx}{\linewidth}{ccc}
		\hline
		\textbf{Label} & \textbf{Meaning}           & \textbf{Data Type} \\ 
		\hline
		NAM            & Name                       & str                \\
		BEZ            & Label                      & str                \\
		BRF            & Lane Width                 & str                \\
		FSZ            & Lane Number                & int                \\
		WDM            & Classification             & str enum           \\
		FTR            & Roadway Separation         & str enum           \\
		STS            & Road Key                   & str                \\
		IBD            & International Significance & str enum           \\ 
		\hline
	\end{tabularx}
\end{table}
Whereas in railways the attributes of the railway shape objects ``AX{\textunderscore}Bahnstrecke'' and ``AX{\textunderscore}Bahnverkehr'' were queried for the following key-value pairs as listed in \cref{tab:railway_attributes_table}. The derived attributes' values from the ATKIS masks are then used for set operations on the clustered above ground LAS layer with an optimally derived empirical threshold of $>65\%$ overlapping of clusters to automate the labelling of class to balance precision and computing speeds  to extract the vegetation, elevated roads, bridges and railway classes. Thresholds below this value would trigger a further step of sorting for the nearest label from cluster centre that annotates the largest sample of cluster points within a buffer zone of 0.75x the mean width in 2D space~\cite{Glira.2022}. The following classes could be identified using the aforementioned approach:

\begin{enumerate}
	\item Vegetation: high and low.
	\item Road: traffic roads with their pavements, footpaths and parking lots.
	\item Railway: trackbeds including sleepers and ballast, with subclasses for rails, posts and cables could be extracted, but were merged together eventually in one class for balancing reasons.
	\item Crossings: bridges over land, water bodies or railway as well as tunnels and either assigned to road or railway class.
	\item Buildings.
	\item Miscellaneous class: ``undefined'' for found objects that fit none of the aforementioned categories, like sparse bird flock in Jena, vehicles in Erfurt, ships and boats in Dresden, transmission towers in Weimar, etc.
\end{enumerate}

\begin{table}[H]
	\centering
	\caption{Relevant railway attributes extracted from the shape of the railway axes shape objects ``AX{\textunderscore}Bahnstrecke'' and ``AX{\textunderscore}Bahnverkehr'' for labelling PCD.}
	\label{tab:railway_attributes_table}
	\newcolumntype{c}{>{\centering\arraybackslash}X}
	\begin{tabularx}{\textwidth}{ccc}
		\hline
		\textbf{Label}	& \textbf{Meaning}	& \textbf{Data Type}\\
		\hline
		BKT & Rail Category                            & str enum \\
		ELK & Electrification                          & str enum \\
		GLS & Number of Tracks                         & str      \\
		NRB & Rail Route Number                        & str      \\
		SPW & Track Gauge                              & str enum \\
		ZUS & Condition                                & str enum \\
		\hline
	\end{tabularx}
\end{table}
For the generation of 2D masks from orthophotos, two methods have been tested: (1) Using overlay set operations with derived geometries from the ATKIS shapes on synthetically generated connected components labelled images, which originate from the Gaussian mixture model fit of the pixels RGB values of the original orthophotos in colour space as features and (2) trials of instance segmentation via inference with zero-shot generalisation from the Segment Anything Model~\cite{Kirillov.2023}. Especially in cases of pale vegetation and buildings with low contrasting coloured roof tops to the surroundings, (2) resulted in better image clusters for annotations.

\subsection{Training and Testing of Point Cloud Segmentation DNN Framework}
The processed dataset included 313 grid cells of 1x1 kilometres from the cities of Erfurt, Jena and Weimar in Thuringia and 236 from the cities of Leipzig and Dresden of 2x2 kilometres in Saxony respectively with a total number of 1,257 cells after quad-splitting the grid cells from Saxony to 1x1 kilometres. Every cell was further quad split with the number of points in each block chosen at 4096 for training with limited GPU resources. The following classes were annotated: buildings, vegetation, water bodies, roads, railway trackbeds and a miscellaneous class for none of the aforementioned categories.

The custom dataset had to be manipulated and converted through an intermediary MMDetection3D format to provide description of the 3D boxes with 600 epochs for training of modified 2DPASS architecture.  The best trained model achieved a semantic segmentation result of 71.48\% for the Mean Intersection over Union (MIoU). Further sensitivity analysis steps to optimise the training hyper-parameters are needed and an increase in number of points per block to 8192 should be investigated to determine whether the segmentation result can yield a higher value for the evaluation metric.
\subsection{Semantic Enrichment for BIM}
This step takes the segmentation results along with the colourised PCD sampled from orthophotos to construct 3D primitive shapes and enrich them with generated context from the derived segmentation and related GIS data or meshes 3D surfaces with texture with a focus on IFC compatibility to be integrated in the information model.
\begin{table}[H] 
	\centering
	\caption{Freely available segmentation output data used within the implementation and their suggested relevant shape representation and entity in IFC.}
	\label{tab:las_ifc_entities_mapping}
	\newcolumntype{c}{>{\centering\arraybackslash}X}
	\begin{tabularx}{\textwidth}{cccc}
		\hline
		\textbf{Segmentation Class}	& \textbf{LAS Class} & \textbf{IFC Product}		& \textbf{IFC Entity}\\
		\hline
		Ground Terrain              & 2           		& Tri/PolygonalFaceSet 			& \textit{IfcSite}     \\
		Overground                  & 20          		& -                   			& \textit{IfcProxy}    \\
		Vegetation                  & 2, 3, 4, 5, 20    & FacetedBrep                	& \textit{IfcProxy}    \\
		Buildings                   & 0, 1, 6, 20    	& Tri/PolygonalFaceSet 			& \textit{IfcBuilding} \\
		Water                       & 2, 9         		& Tri/PolygonalFaceSet 			& \textit{IfcProxy}    \\
		Roads                       & 2, 11        		& Tri/PolygonalFaceSet 			& \textit{IfcPavement} \\
		Railway Body                & 2, 10        		& FacetedBrep 					& \textit{IfcProxy}    \\
		Track Alignment				& 2, 10        		& PositioningElement	 		& \textit{IfcAlignment}\\
		Miscellaneous               & 0, 1, 2       	& Tri/PolygonalFaceSet 			& \textit{IfcProxy}    \\
		Unclassified                & 0, 1         		& FacetedBrep                	& \textit{IfcProxy}    \\
		\hline
	\end{tabularx}
\end{table}
\subsubsection{Constructing IFC-Alignment Objects}
In order to create not only geometrically but also semantically proper \textit{IfcAlignment} entities, the segment type needs to be assigned to the respective alignment points that originate from the semantic segmentation. Hence, it is not sufficient to only assign the affiliation as `alignment', but also the segment type, such as line, curve or clothoid, needs to be assigned to every point. 

In a similar approach to Lin et al.~\cite{Lin.52020095222009}, the prediction mask’s boundaries containing railway tracks and connecting bridges could be isolated in a projected 2.5D raster using the Python Packages Geopandas, Shapely, Rasterio and CloudCompare in a background process. This served as the foreground to estimate the centreline of the tracks through estimation of a normalised euclidean distance transformation and local maxima extraction with a maximum search radius of the minimum nearest neighbour value. In case of branching or merging of the elongated stripes, cleaning for noise and further processing steps are necessary in order to predict instances properly.

The stripes can be split by skeletonising the boundary shape of the foreground and Hit-or-Miss morphological operations to identify junctions and points to split them into separate segments. Using any of the available line decimation algorithms like Rahmer-Douglas-Peuker or Visvalingam-Whyatt can then estimate a simplified polyline of the split skeleton segments similar to the approach implemented in~\cite{Alabassy.2022}, which can be smoothed out using curve fitting with polynomials. The estimated smoothed curve points could be used to identify their closest points belonging to the actual 2.5D raster and retrieve their original height value using their identifiers.

In the \textit{IfcRail} domain, the construction of railway alignments requires many different entities due to the parametric nature of railway planning. Within the IFC 4.3.x specification, horizontal alignments can be one of four different segment types: lines, circular arcs, linear curvatures like clothoids or nonlinear curvatures like Helmert Curves~\cite{buildingSMART.2023b}. All horizontal segments are, among other parameters, defined by their StartPoint (a cartesian point of x- and y-coordinates), a StartDirection (expressed as an angle) and their SegmentLength, which leads overall to the next so-called station of the alignment segment. In case of curves and clothoids, further parameters, such as the radius, are necessary to construct the arithmetical geometric elements. Those elements form the horizontal alignment on a plane. To display also the horizontal course of the height, the construction of an \textit{IfcGradientCurve} is necessary. The final alignment must always include both the vertical and horizontal alignment.

The segmented points of the railway track do not yet contain the information about their segment affiliation. Therefore, fitting algorithms are used to identify different segment types. Once the coordinates can be assigned to a segment type (e.g., \textit{IfcLine}, \textit{IfcClothoid} and \textit{IfcCurveSegment}), the \textit{IfcAlignment} objects can be constructed. In the end, not all coordinates are necessary to define a segment. By identifying the start and end point of a segment, the other points, which have been assigned to the segment, can be used in order to derive other necessary parameters, such as radius, tangent direction or a segment length.

To estimate the segment types and their parameters, a 2.5D raster of the segmented inference for the railway tracks as a binary image can be generated. This raster can be thresholded into a binary image and further processed as an image with a morphological skeletonisation step to extract the medial axis of the railway track instances. Based on the skeletonisation method used, further processing steps to clean and smooth the axes may be required. Afterwards, the Hough transformation can be used on the skeletonised 2.5D raster to detect straight lines and circular segments. The remaining parts of the medial axes can be fitted with clothoids by interpolating between detected end points and/or junctions detected using binary Hit-or-Miss transform in a plane with assigned tangent directions derived from the oriented gradient map of the euclidean distance transform via $G^1$ Hermite interpolation.

\subsubsection{Colour Sampling and Texturing}
\begin{figure}[H]
	\centering
	\includegraphics[width=0.8\linewidth]{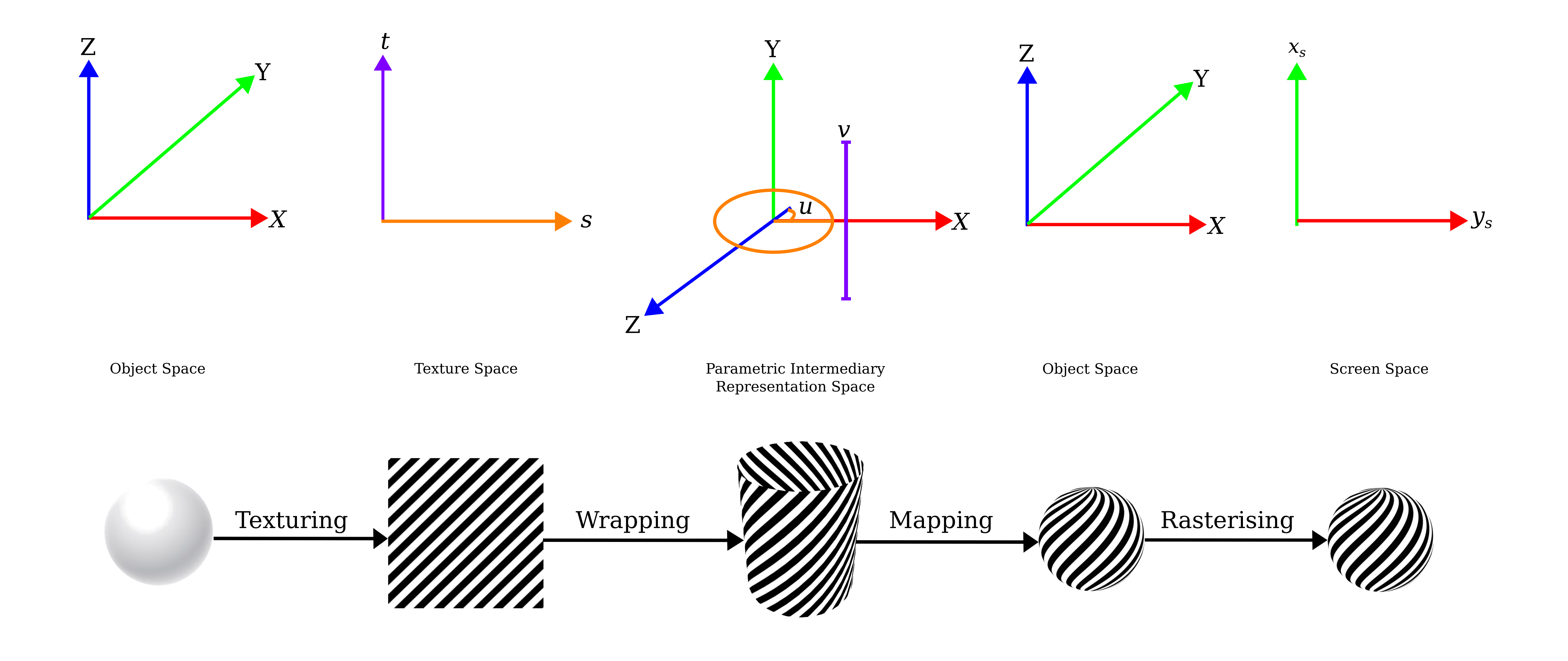}
	\caption{Process for texture mapping through parameterised intermediary geometry.}
	\label{fig:texturing_explained}
\end{figure}

As a single colour value alone is sometimes insufficient to model realistic appearances or provide information, mapping a texture to a 2D or 3D surface like gift wrapping is a long established process in animation and graphics fields but it is rarely adopted in BIM related application, alas in research specific context of damage modelling where most of the case studies published have been focused. The different spaces considered when unwrapping and mapping texture to a 3D or 2D surface in 3D are shown in~\cref{fig:texturing_explained}. The texture mapping to the object space could be directly by defining a parametric representation of a surface (e.g., sphere) to map U,V to S,T within the interval [0,1]. 

By defining the geometric representation of the shape mathematically the (U, V) of each vertex can be matched to its correspondent (S, T). This could be achieved by directly defining the mathematical equations for a surface or by, manually unwrapping a 3D shape into a 2D surface using UV-mapping and editing specialised software or by using an intermediate geometry for projecting the texture (e.g., a cylinder) for projection when a discrete or symbolic definition of the shape cannot be easily or directly formulated, especially in free-formed complicated 3D shapes. Each of the aforementioned options has its own advantages and limitations depending on the parameterisation constraints and method used as well as the texture pattern itself.

\begin{figure}[H]
	\centering
	\begin{subfigure}{0.3\linewidth}
		\includegraphics[width=\linewidth]{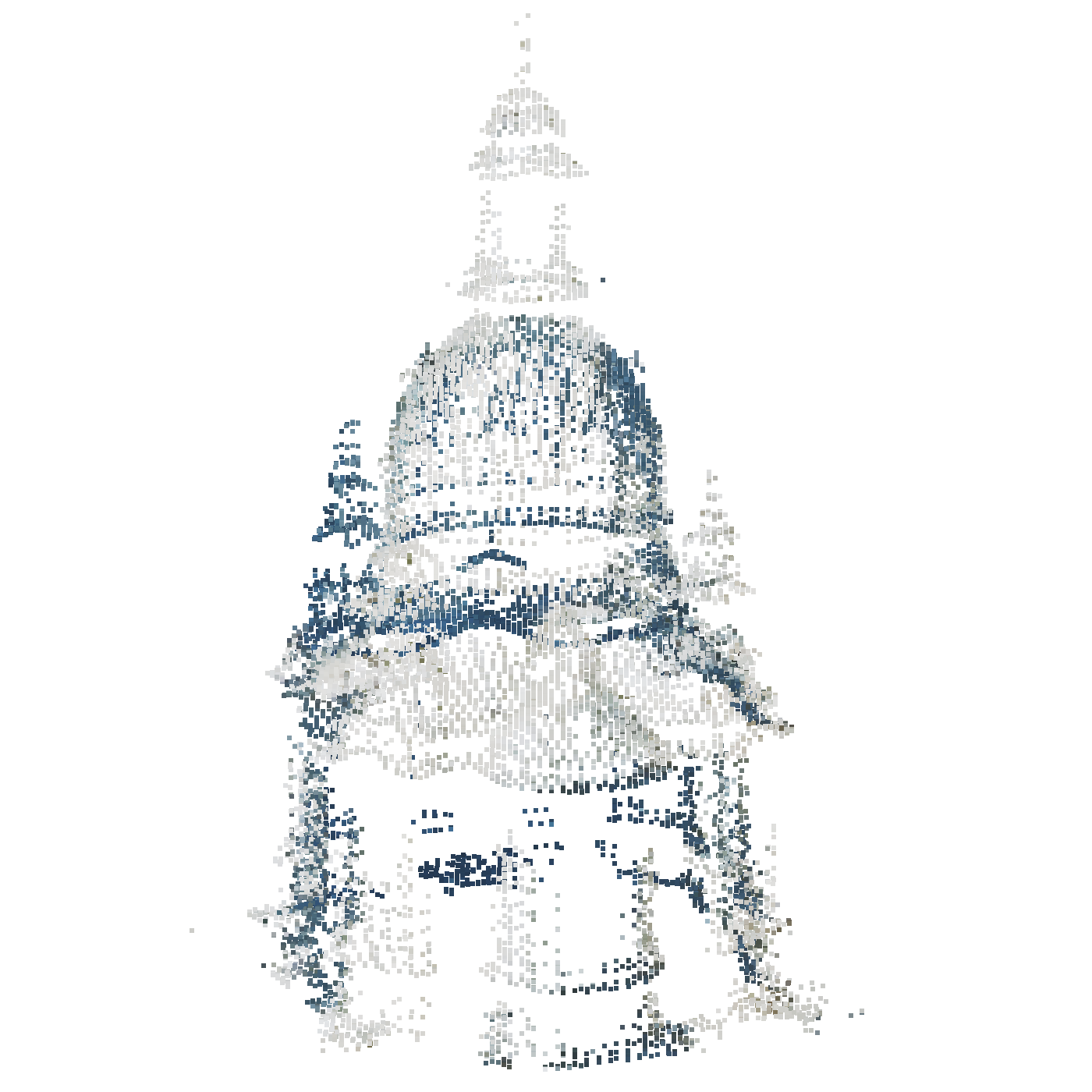}
		\caption{}
	\end{subfigure}
	\begin{subfigure}{0.3\linewidth}
		\includegraphics[width=\linewidth]{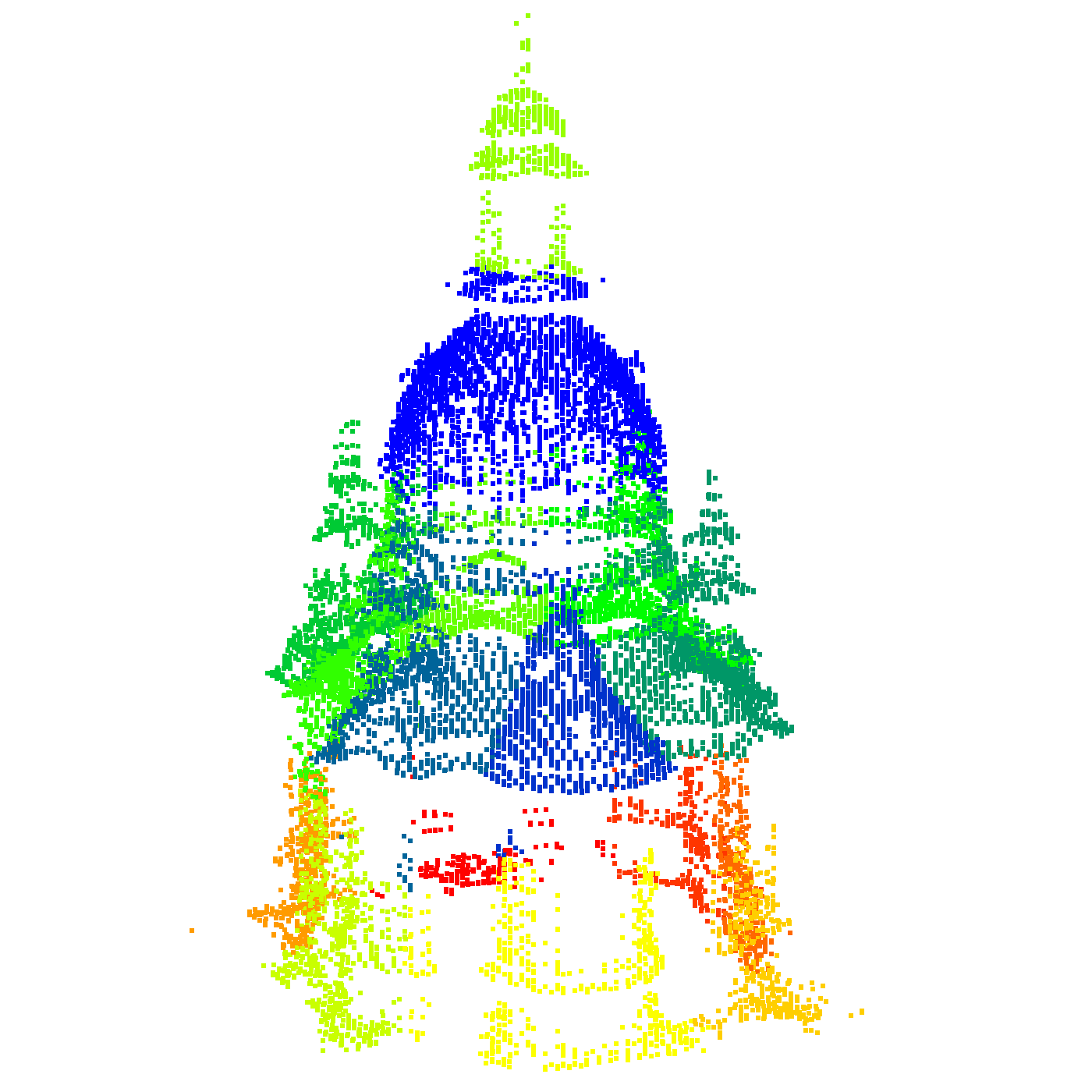}
		\caption{}
	\end{subfigure}
	\begin{subfigure}{0.3\linewidth}
		\includegraphics[width=\linewidth]{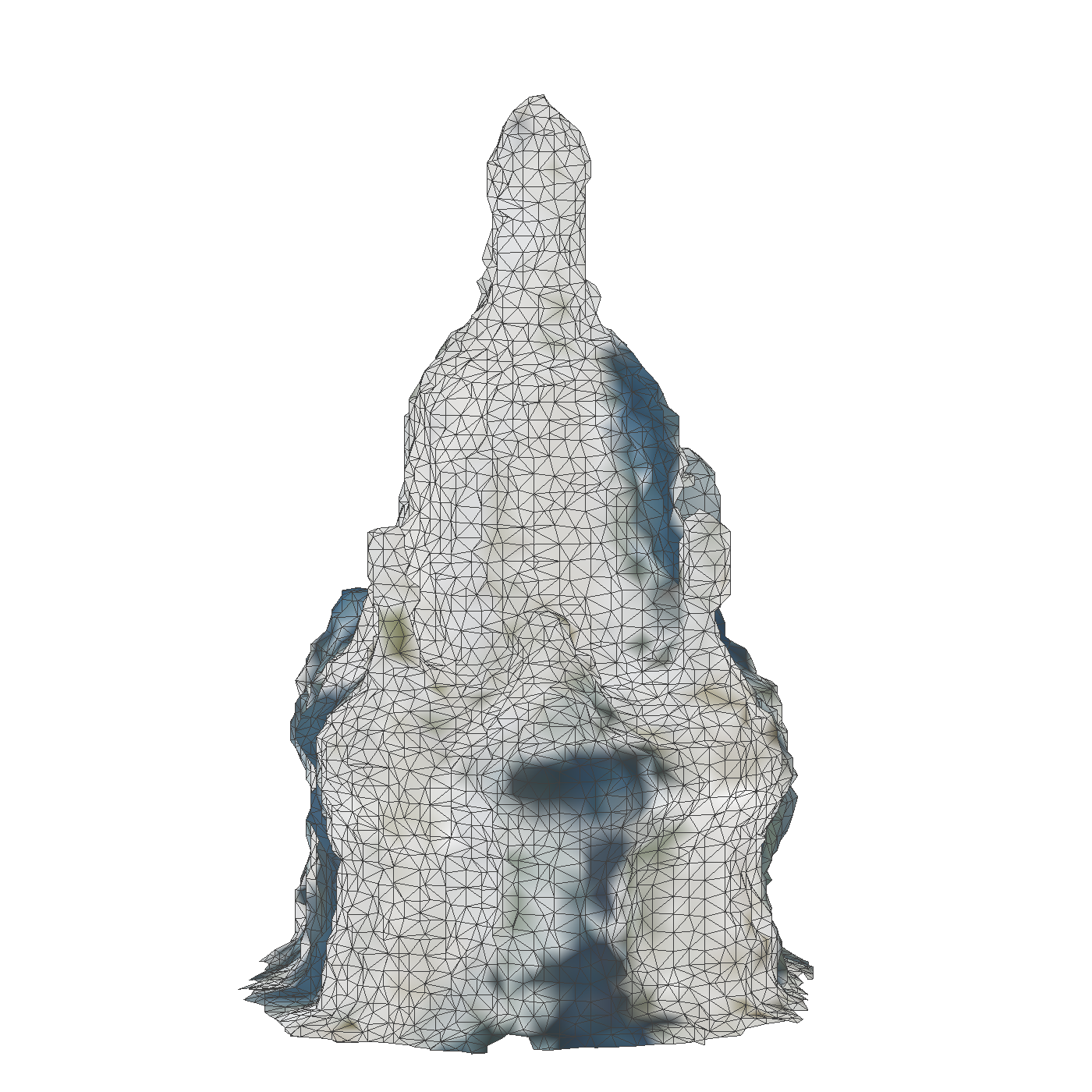} 
		\caption{}
	\end{subfigure}\\
	\begin{subfigure}{0.3\linewidth}
		\includegraphics[width=\linewidth]{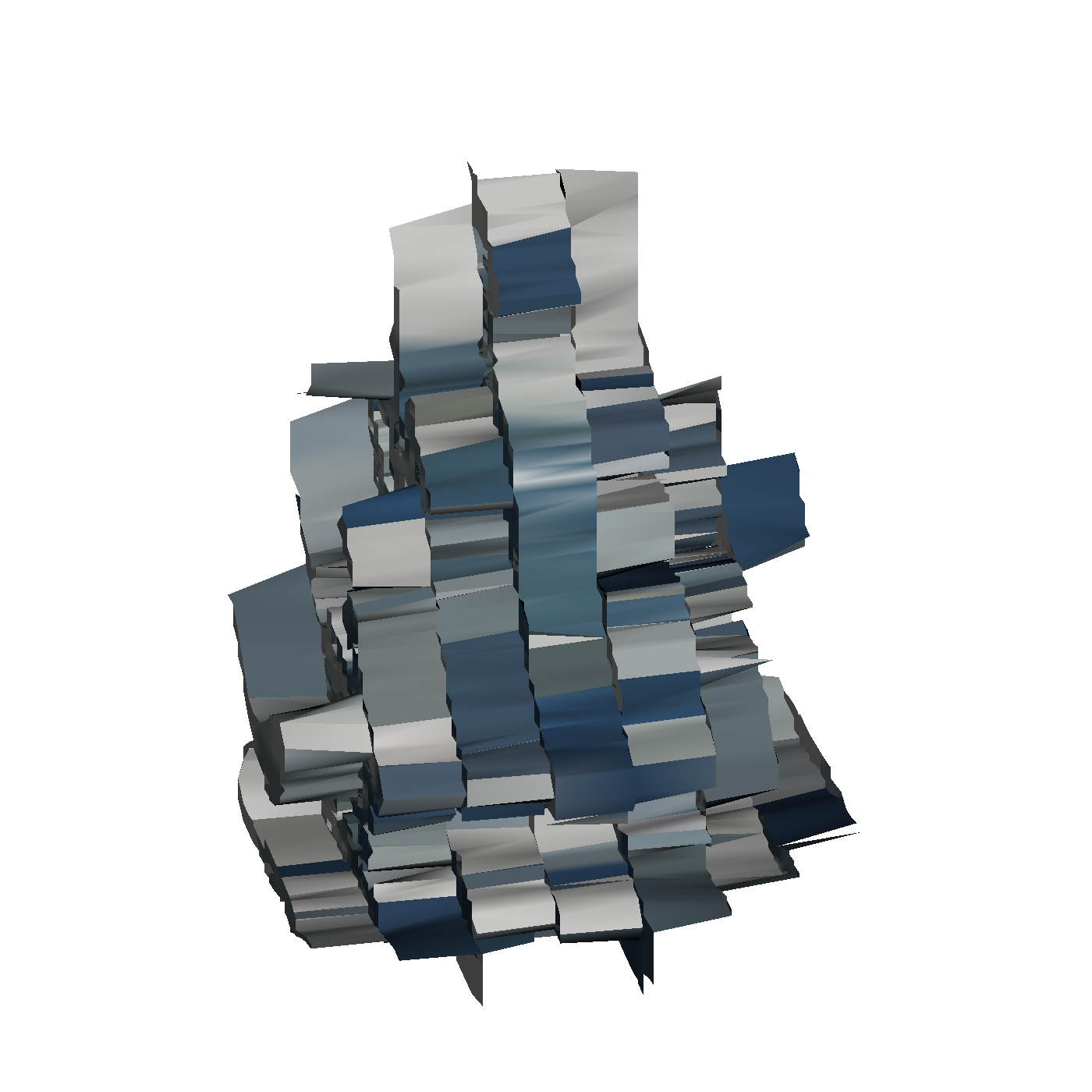} 
		\caption{}
	\end{subfigure}
	\begin{subfigure}{0.3\linewidth}
		\includegraphics[width=\linewidth]{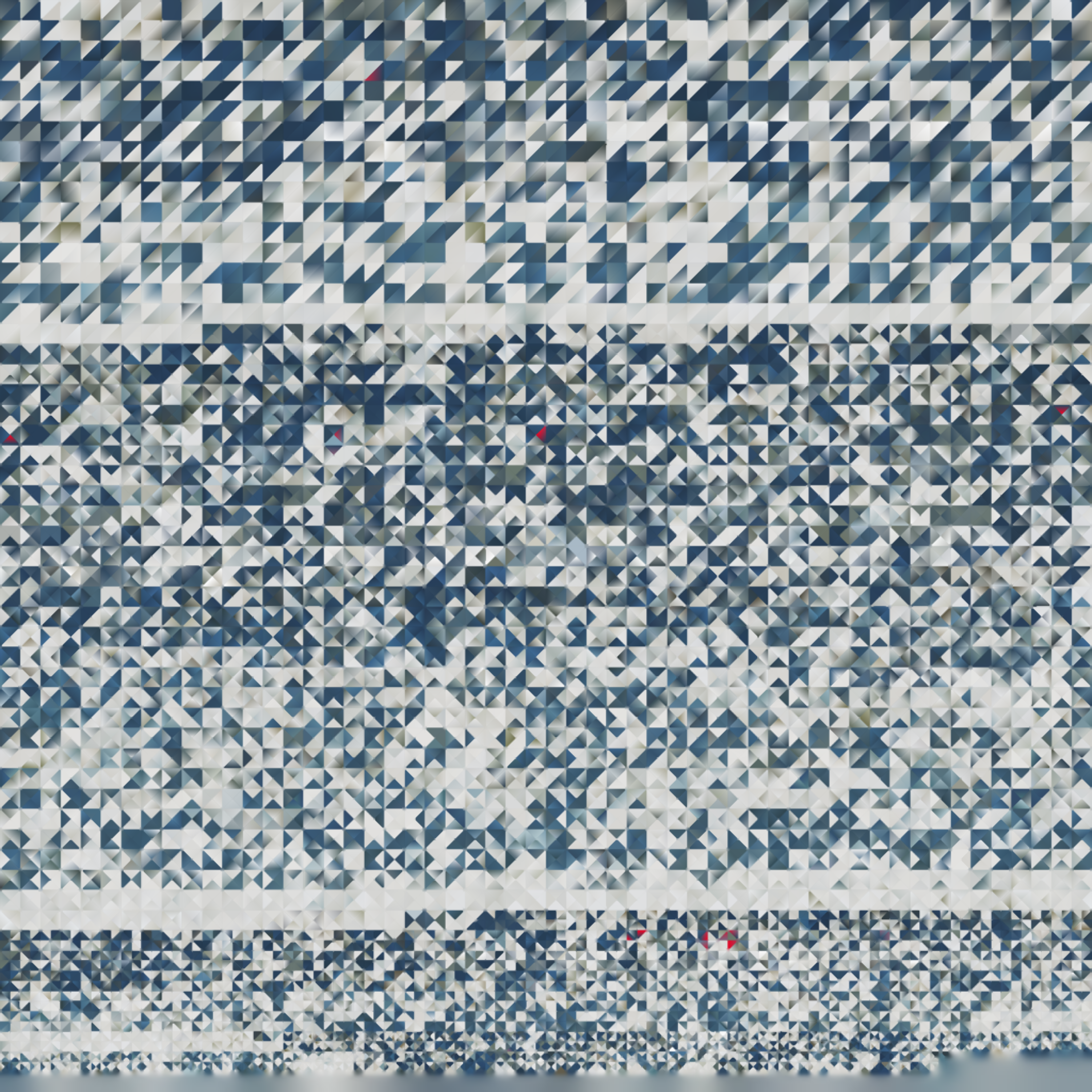}
		\caption{}
	\end{subfigure}
	\begin{subfigure}{0.3\linewidth}
		\includegraphics[width=\linewidth]{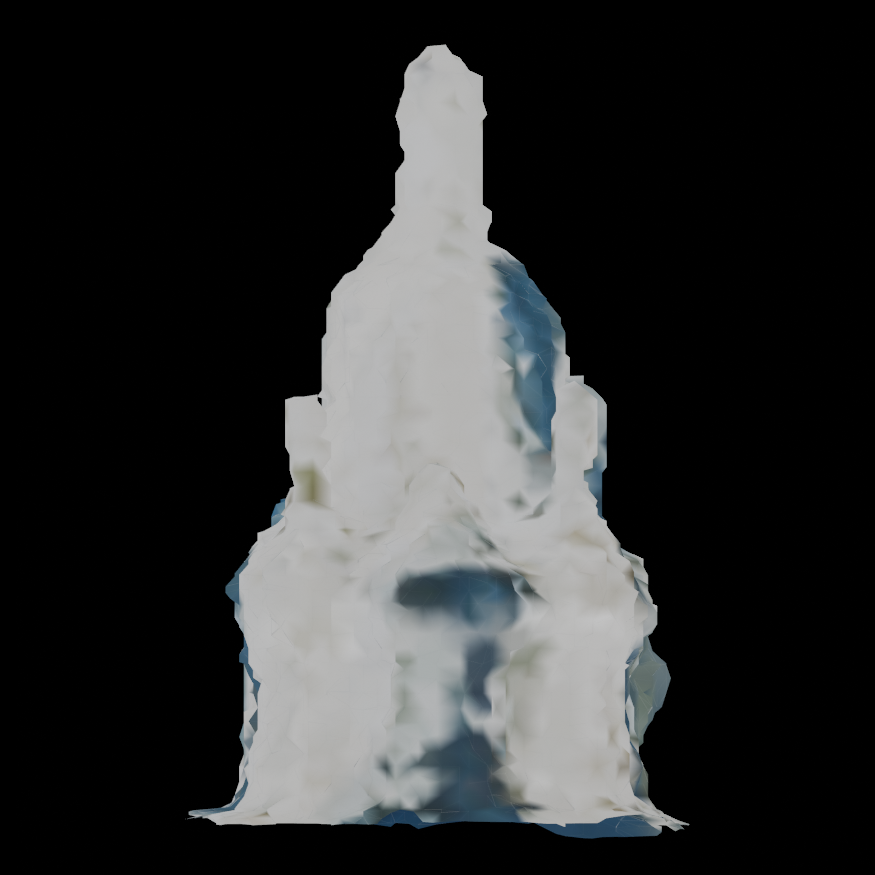}
		\caption{}
	\end{subfigure}
	\caption{Overview of the 3D reconstruction process of a building instance; (a) 3D view of a point cloud instance of the Frauenkirche in Dresden, (b) Randomly coloured clusters of the point cloud instance via mean shift clustering algorithm, (c) Final triangle meshed surface of the point cloud after iterative subsampling, ball pivoting surface reconstruction and triple axial dust accumulation simulation finalised with a screened Poisson surface reconstruction, (d) 3D view of the intermediary parametrised geometry to the UV-mapped mesh surface in texture space, (e) Final texture atlas of the triangle meshed surface of the point cloud as an RGB colour space image in PNG format. (f) IFC textured model viewed with BlenderBIM.}
	\label{fig:Frauenkirche_pcd_surface_reconstruction}
\end{figure}

A Python script has been developed to generate a UV-mapped textured 2-manifold mesh. The Trimesh library is used to create a mesh from the point cloud using the Poisson surface reconstruction. Downsampling and normals estimation are performed using Open3d~\cite{Zhou.2018}, where vertices with low density are removed from the mesh to reduce its size. A texture atlas is automatically generated for UV-mapping, where texture coordinates at edges are generated automatically into a collection of parametrised texture maps compiled together into an atlas~\cite{Levy.2002,Sorkine.,Sander.2001} as Portable Network Graphics (PNG) image. The UV-mapped mesh is regenerated using Pymeshlab with vertex colouring information transferred to a texture and saved as an .obj linked to a texture atlas as a .png image defined through a material .mtl file. \Cref{fig:Frauenkirche_pcd_surface_reconstruction} displays the process of 3D reconstructing a textured mesh from the building instance vertices of the Church of our Lady (i.e., Frauenkirche).

The use of the \textit{IfcIndexedTriangleTextureMap} entity to model a texture on a triangulated mesh as \textit{IFCTriangulatedFaceSet} has been successfully attempted a few times. It shows great potential to model texture-relevant information in IFC within its initial concept of information modelling for damages in bridges and beyond to other infrastructure domains. However, the inclusion of coloured textures has not been widely adopted due to the difficulties of mapping textures to 3D models automatically and visualising them in IFC viewers correctly.

The Blender software with the BlenderBIM addon proved in this study to be the most suitable for this task. The future expansion of the IFC schema to include new entities like \textit{IfcIndexedPolygonalTextureMap} to provide the means to map 2D textures on quad-meshes or as an \textit{IfcPolygonalFaceSet}~\cite{buildingSMART.2023}. \Cref{tab:las_ifc_entities_mapping} lists the possible segmentation results via inference and the corresponding IFC entities suitable for their integration into BIM.
	\section{Case Study}
\label{}
\begin{figure}[H]
	\centering
	\begin{subfigure}{0.8\linewidth}
		\centering
		\includegraphics[width=\linewidth]{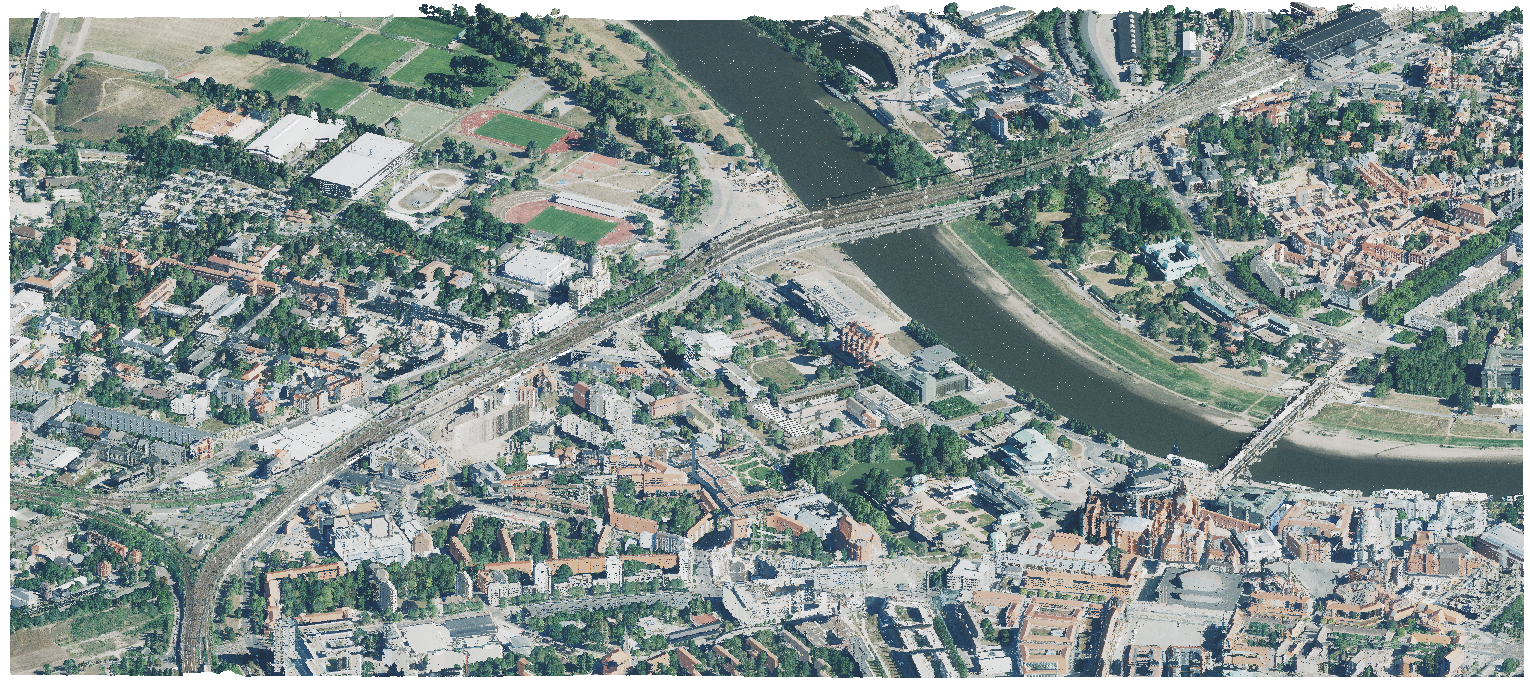}
		\caption{}
	\end{subfigure}
	\begin{subfigure}{0.4\linewidth}
		\includegraphics[width=\linewidth]{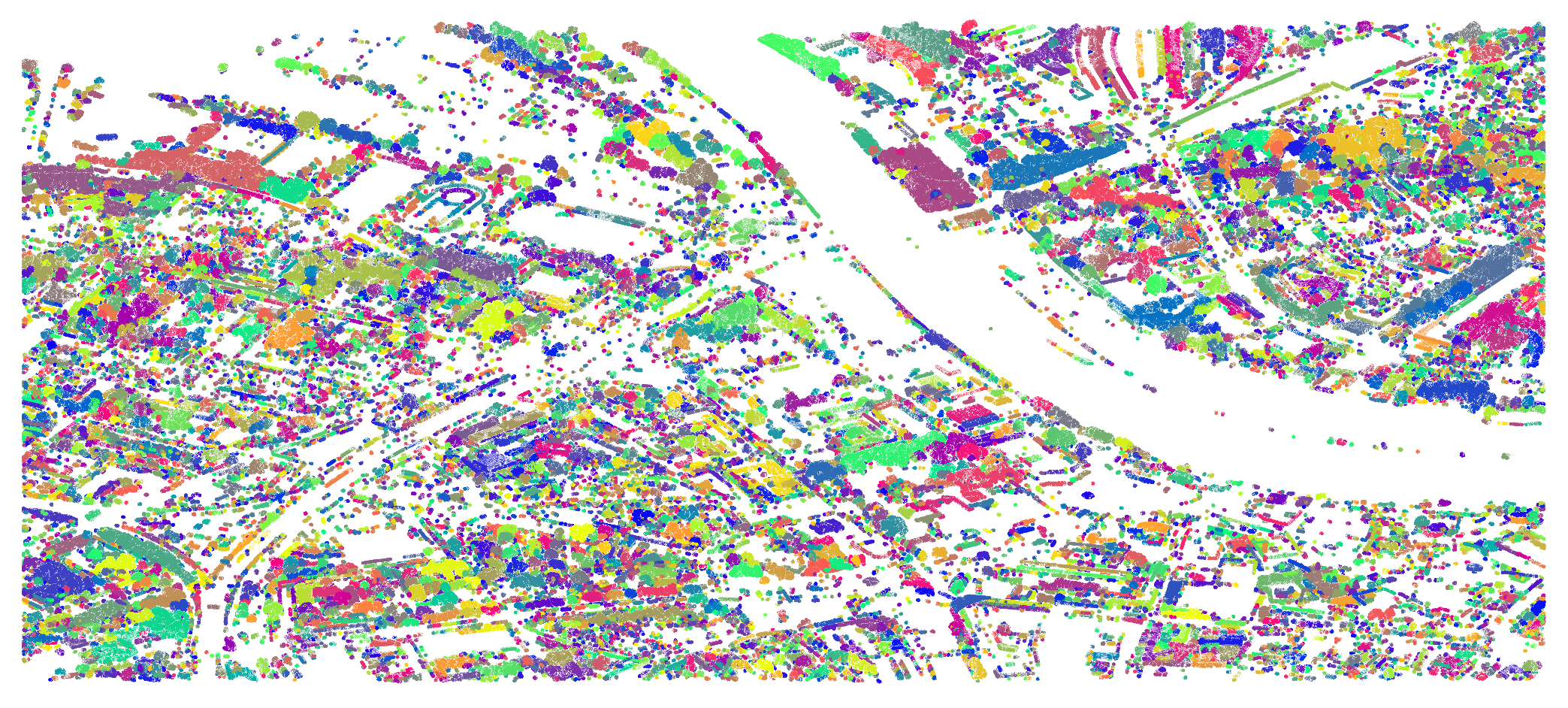}
		\caption{}
	\end{subfigure}
	\begin{subfigure}{0.4\linewidth}
		\includegraphics[width=\linewidth]{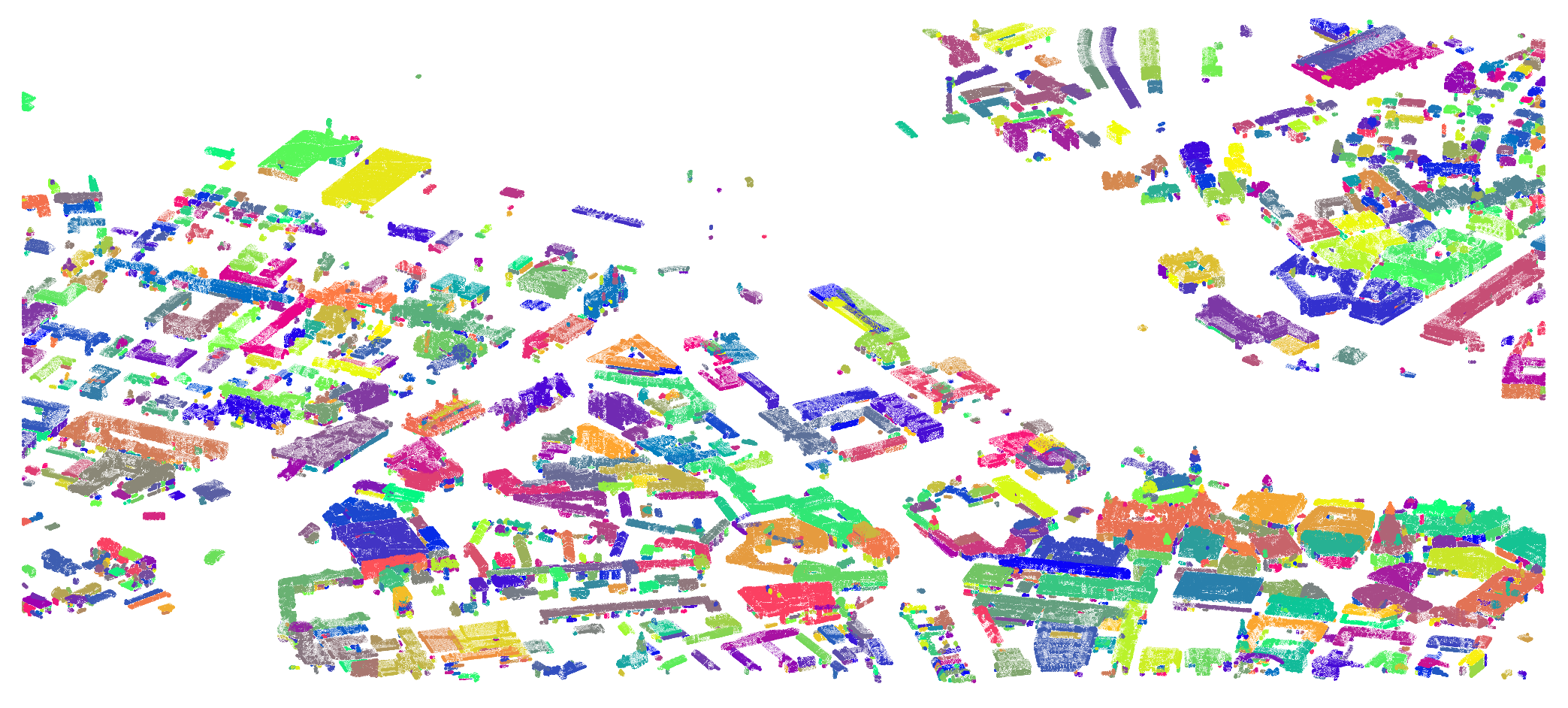}
		\caption{}
	\end{subfigure}
	\begin{subfigure}{0.4\linewidth}
		\includegraphics[width=\linewidth]{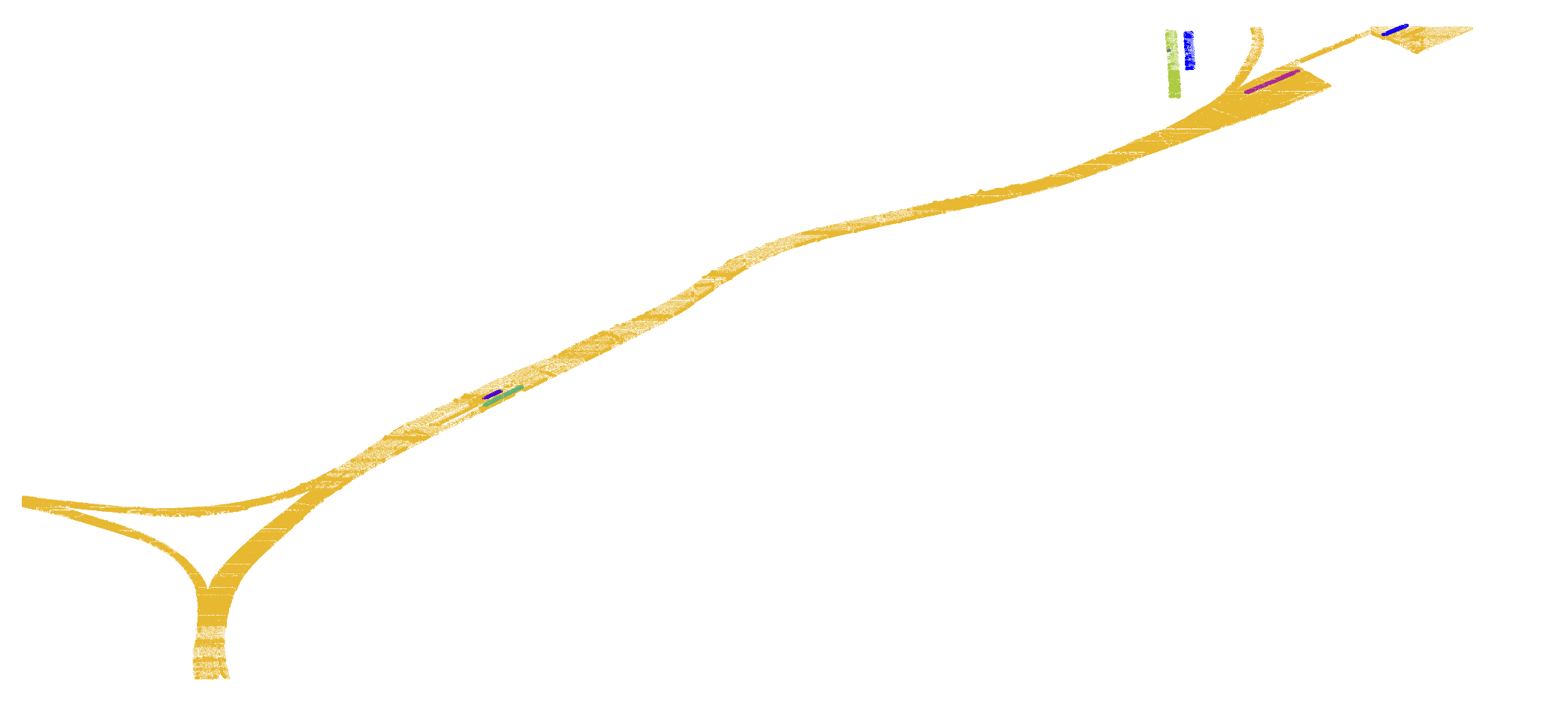}
		\caption{}
	\end{subfigure}
	\begin{subfigure}{0.4\linewidth}
		\includegraphics[width=\linewidth]{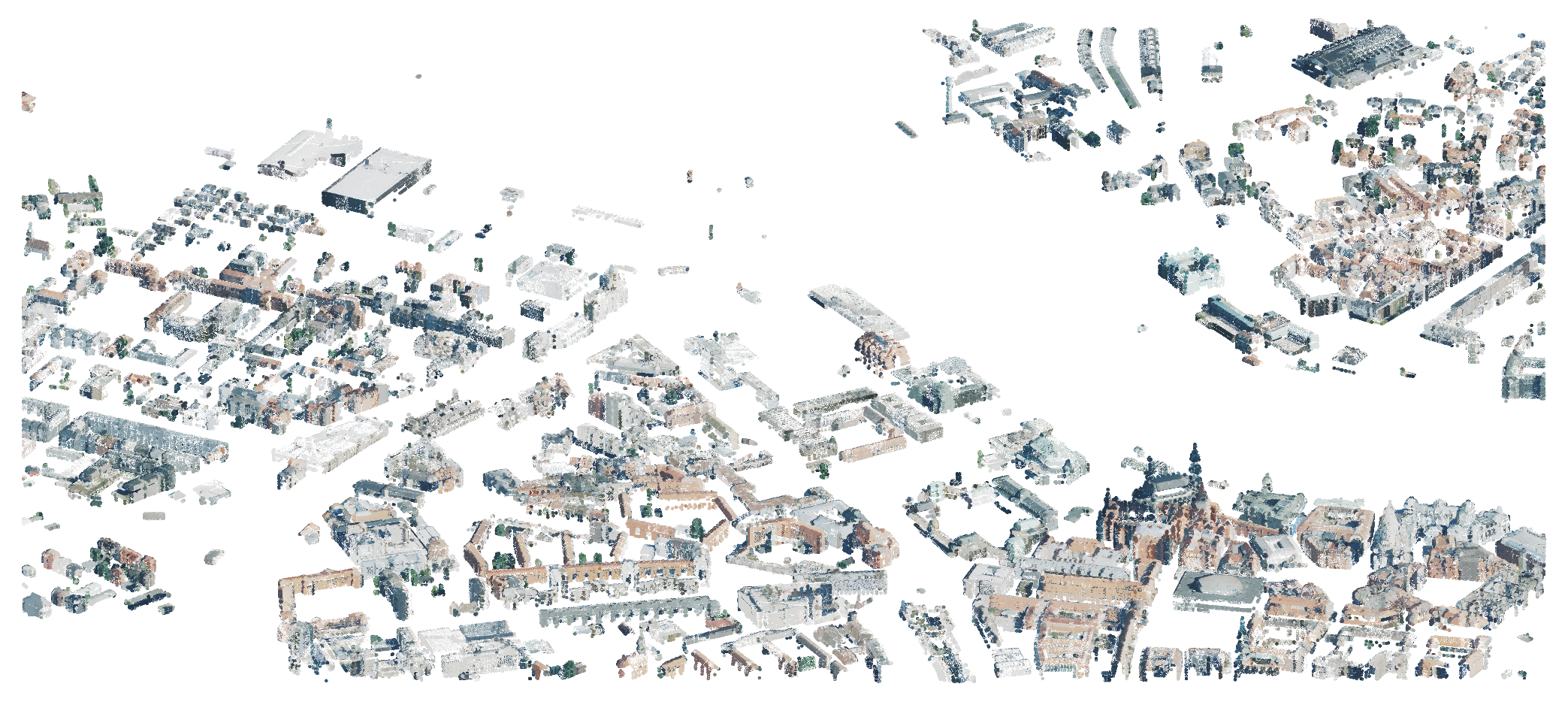}
		\caption{}
	\end{subfigure}
	\caption{Point cloud semantic segmentation; (a) Colourised Aerial LiDAR scan, (b) Segmented vegetation via inference followed by connected component clustering, (c) Segmented buildings via inference followed by connected component clustering, (d) Segmented railway trackbeds via inference followed by connected component clustering, (e) Sampled colours to the clustered instances of segmented buildings.}
	\label{fig:PCD Segmentation}
\end{figure}

Grid cell number ``33410\textunderscore 5656\textunderscore2\textunderscore sn'' in the city of Dresden, Saxony from the test set is used for inference. \Cref{fig:PCD Segmentation} showcases the output result for vegetation, buildings and railway trackbed. A colourised LiDAR scan of the input sampled from its related orthophoto is used to add colours into the segmented vertices. 
\subsection{Recreating Horizontal and Vertical Alignment}
\begin{figure}[H]
	\centering
	\begin{subfigure}{0.3\linewidth}
		\includegraphics[width=\linewidth]{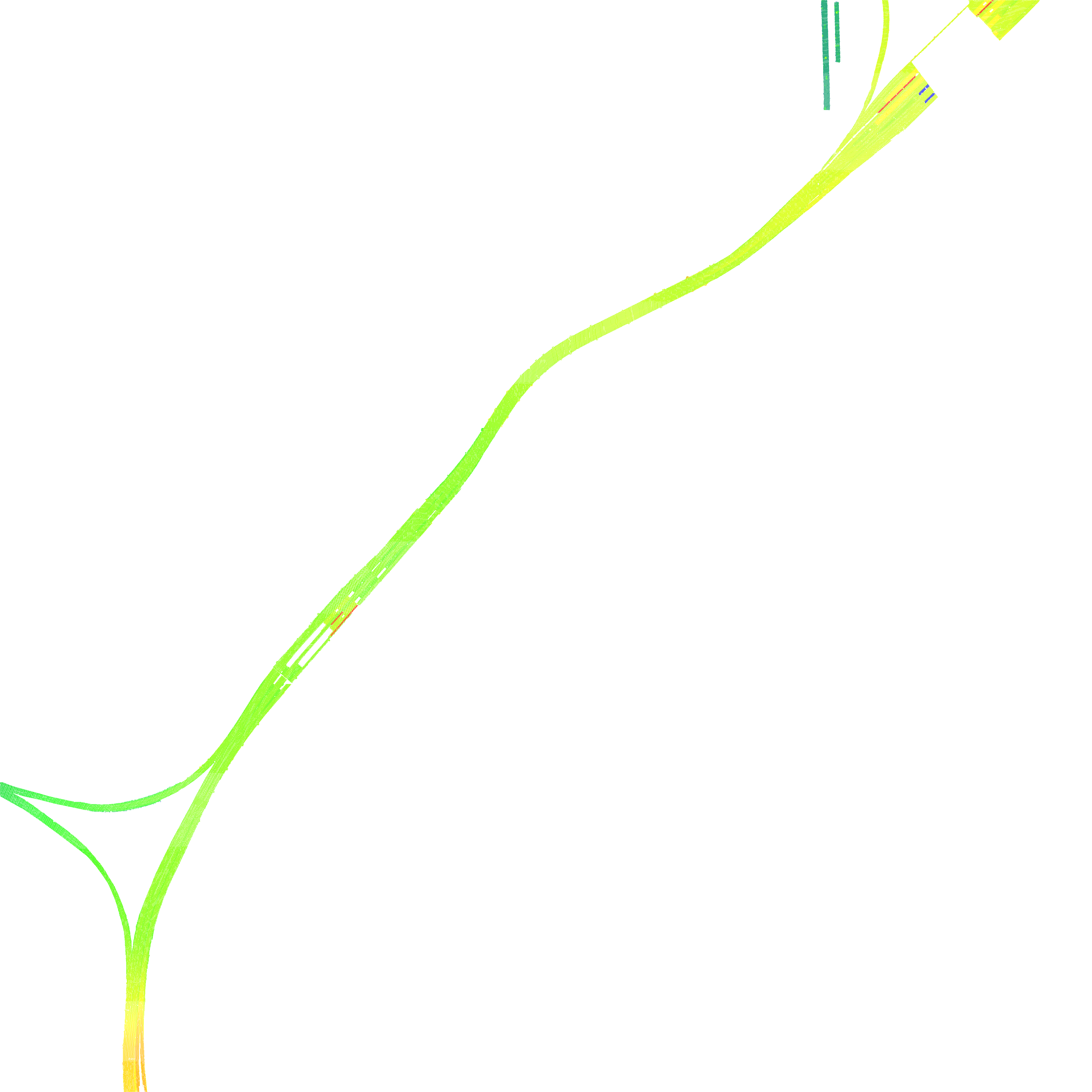}
		\caption{}
	\end{subfigure}
	\begin{subfigure}{0.3\linewidth}
		\includegraphics[width=\linewidth]{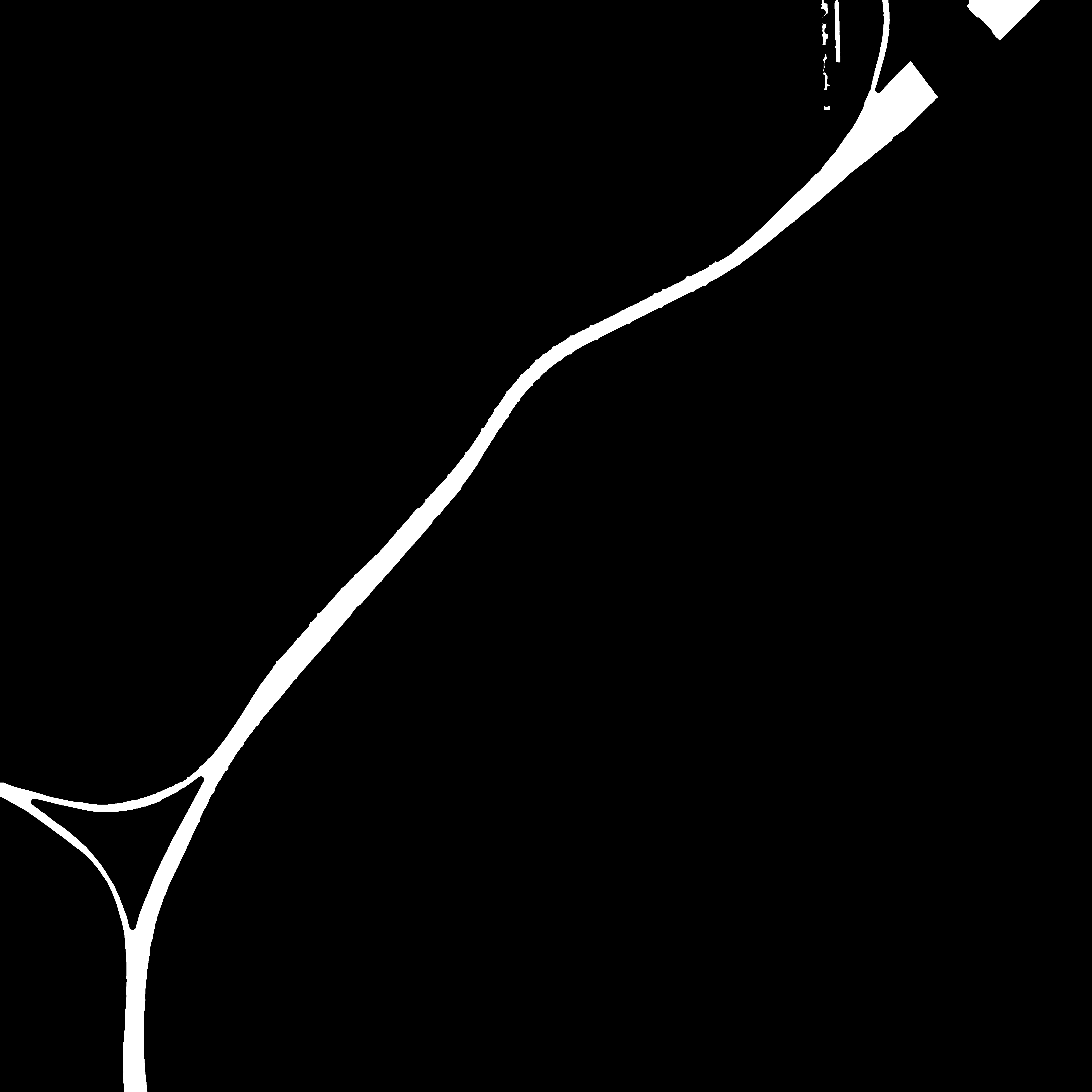}
		\caption{}
	\end{subfigure}
	\begin{subfigure}{0.3\linewidth}
		\includegraphics[width=\linewidth]{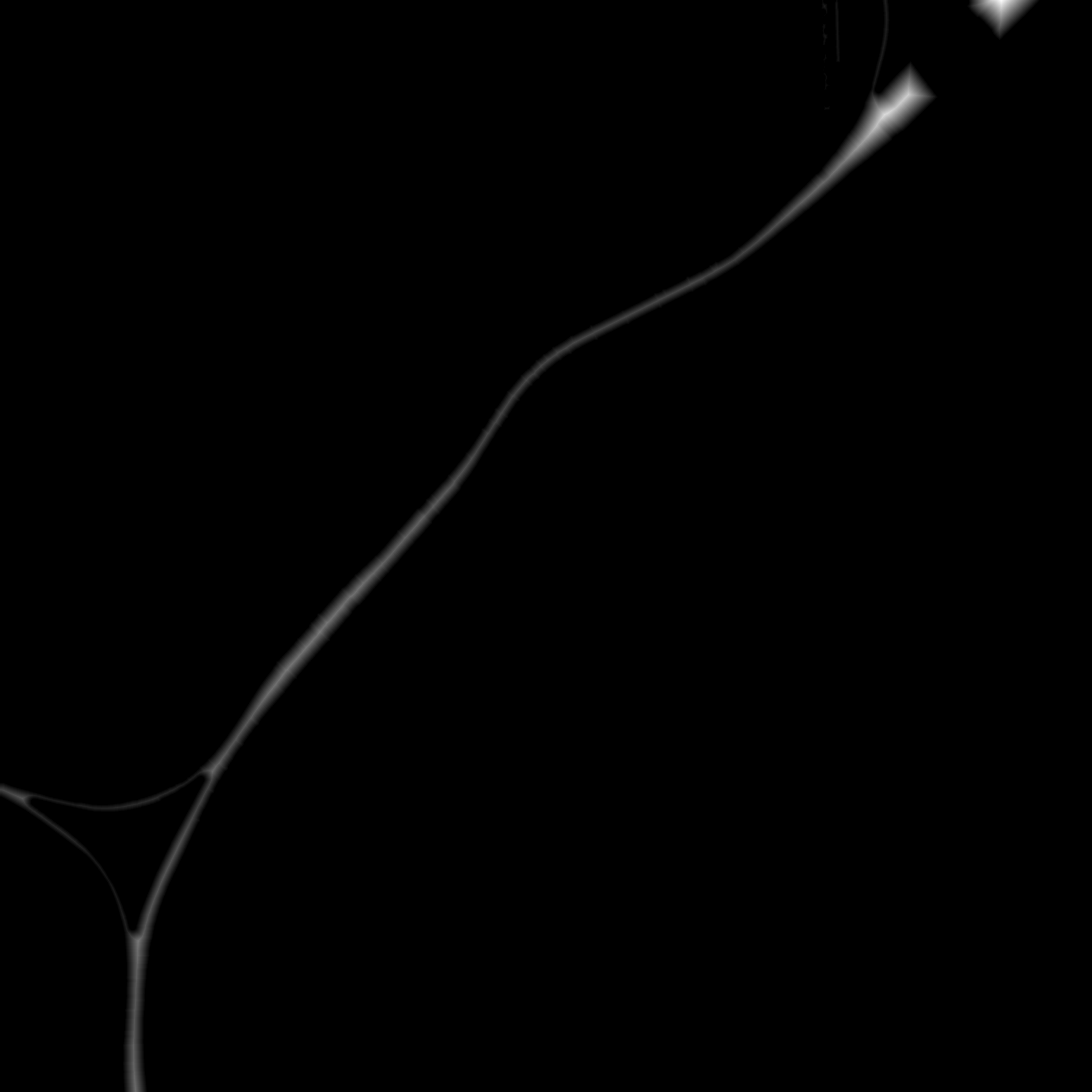} 
		\caption{}
	\end{subfigure}
	\begin{subfigure}{0.3\linewidth}
		\includegraphics[width=\linewidth]{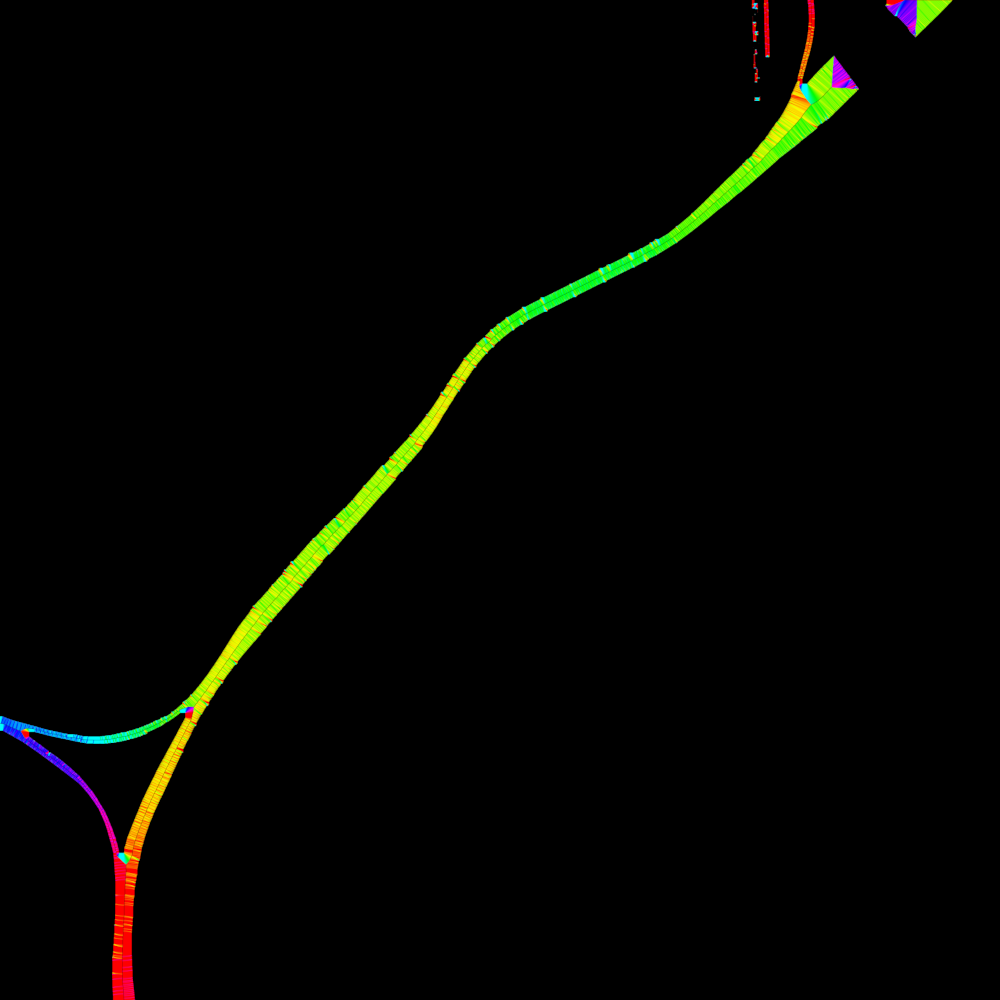}
		\caption{}
	\end{subfigure}
	\begin{subfigure}{0.3\linewidth}
		\includegraphics[width=\linewidth]{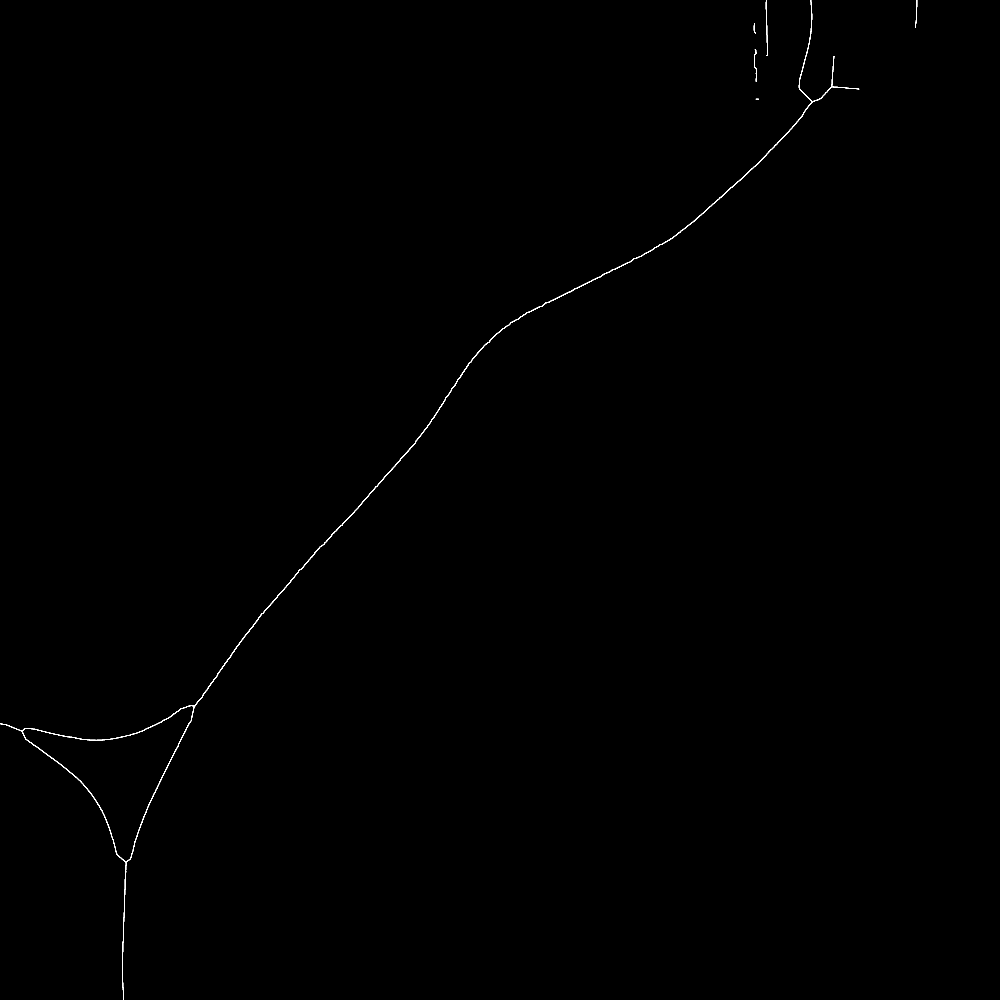}
		\caption{}
	\end{subfigure}
	\begin{subfigure}{0.3\linewidth}
		\includegraphics[width=\linewidth]{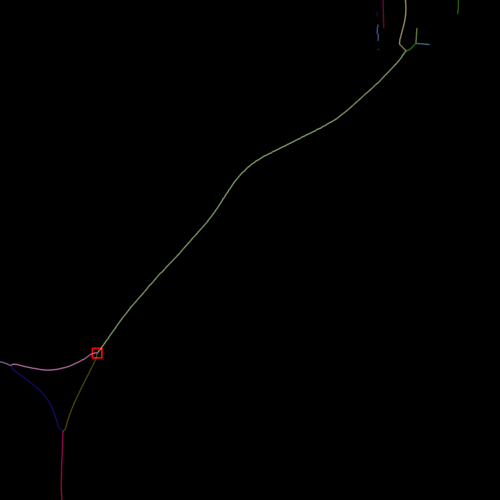}
		\caption{}
	\end{subfigure}
	\begin{subfigure}{0.45\linewidth}
		\includegraphics[width=\linewidth]{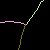} 
		\caption{}
	\end{subfigure}
	\begin{subfigure}{0.45\linewidth}
		\includegraphics[width=\linewidth]{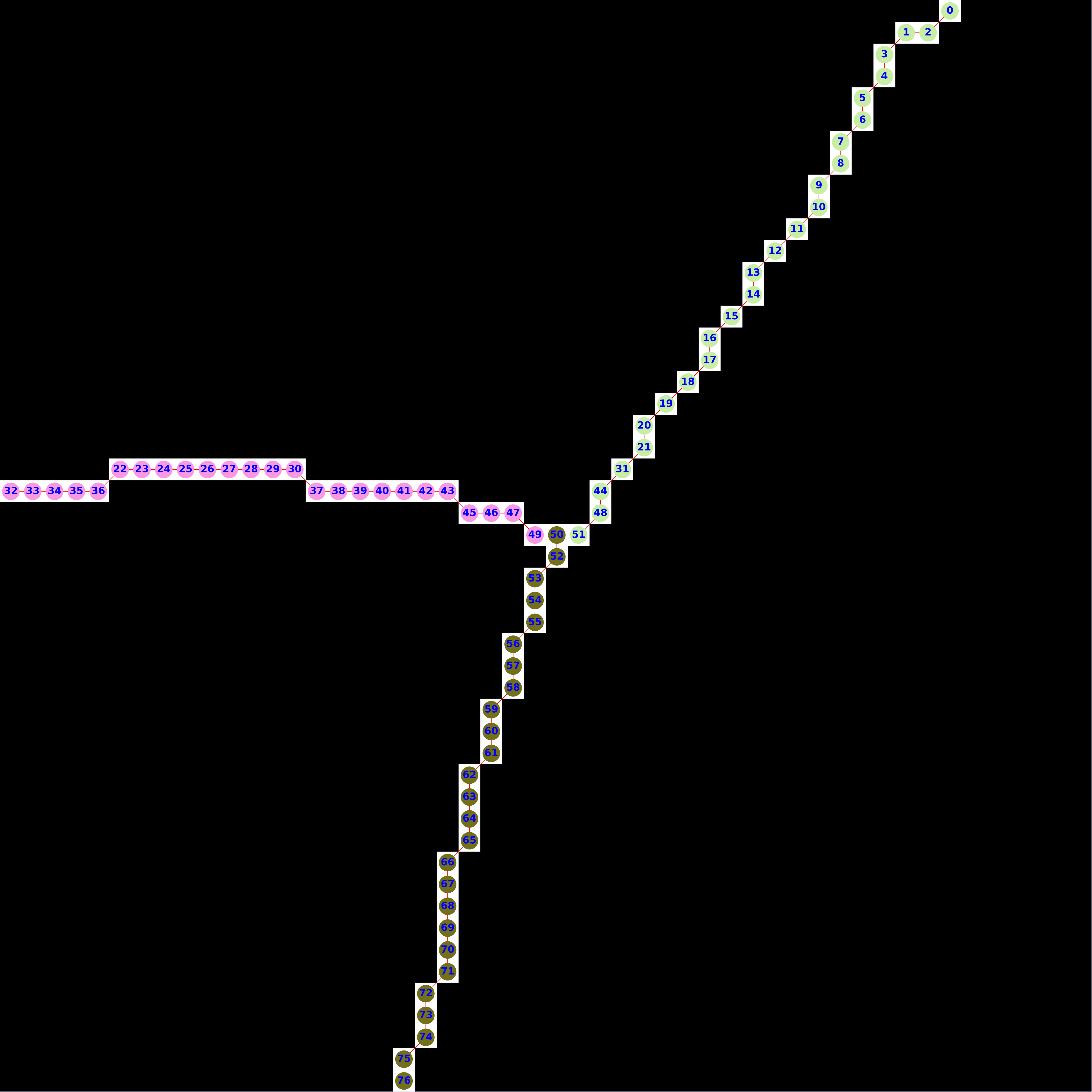}
		\caption{}
	\end{subfigure}
	\caption{Fitting alignment segments to track-bed points. Sub-figures (e) and (f) have been drastically downscaled with cubic interpolation and manipulated by increasing the exposure to allow for better visibility of single pixel thick morphological skeleton in the manuscript; (a) 2.5D Raster image, (b) Binary raster image after opening and closing processing steps, (c) Euclidean Distance Transform map of the binary raster, (d) HSV colour representation of the orientation capped at 180 degrees and normalised magnitude of gradients to the euclidean distance transform map, (e) Skeletonised morphology of the binary image, (f) Labelled segments of the split skeleton assigned to random colours using minimum spanning tree of branching clusters to identify junctions~\cite{NunezIglesias.2018}, (g) A zoomed-in detail of the red bordered region marked in (f) for demonstration, (h) Labelled nodes of the Minimum Spanning Tree for the skeleton pixel branches with a junction node correctly identified at label 50.}
	\label{fig:IFCAlignment}
\end{figure}

The first use-case is demonstrated by generating an estimation of the segmented track-bed of the railway to roughly estimate its alignment. A 2.5D raster of 5000x5000 pixels resolution is generated to facilitate the estimation through classical image processing methods as shown in \cref{fig:IFCAlignment}. First, the image is thresholded to generate a binary image and cleaned from noise and thin elements with opening and closing. Second, the euclidean distance transform (EDT) is calculated to estimate the orientation and magnitude of the gradients of the EDT map. 

Third, the gradient orientation is used as a prior guess of the vector normal to the tangent direction along the median axis. Fourth, the axis itself for the foreground is extracted using morphological skeletonisation and further processed with smoothing and cleaning of noise and small features below 2 metres (i.e., $\approx 5$ pixels). Fifth, the skeleton is split into separate labelled entities by detecting junction and endpoints, removing the junctions temporarily to disconnect the branches of the skeleton from each other to allow for connected component labelling (CCL) then reassemble them to their connected labels. Sixth, the line segments are extracted using the probabilistic Hough transform. 

\begin{figure}[H]
	\centering
	\begin{subfigure}{0.3\linewidth}
		\includegraphics[width=\linewidth]{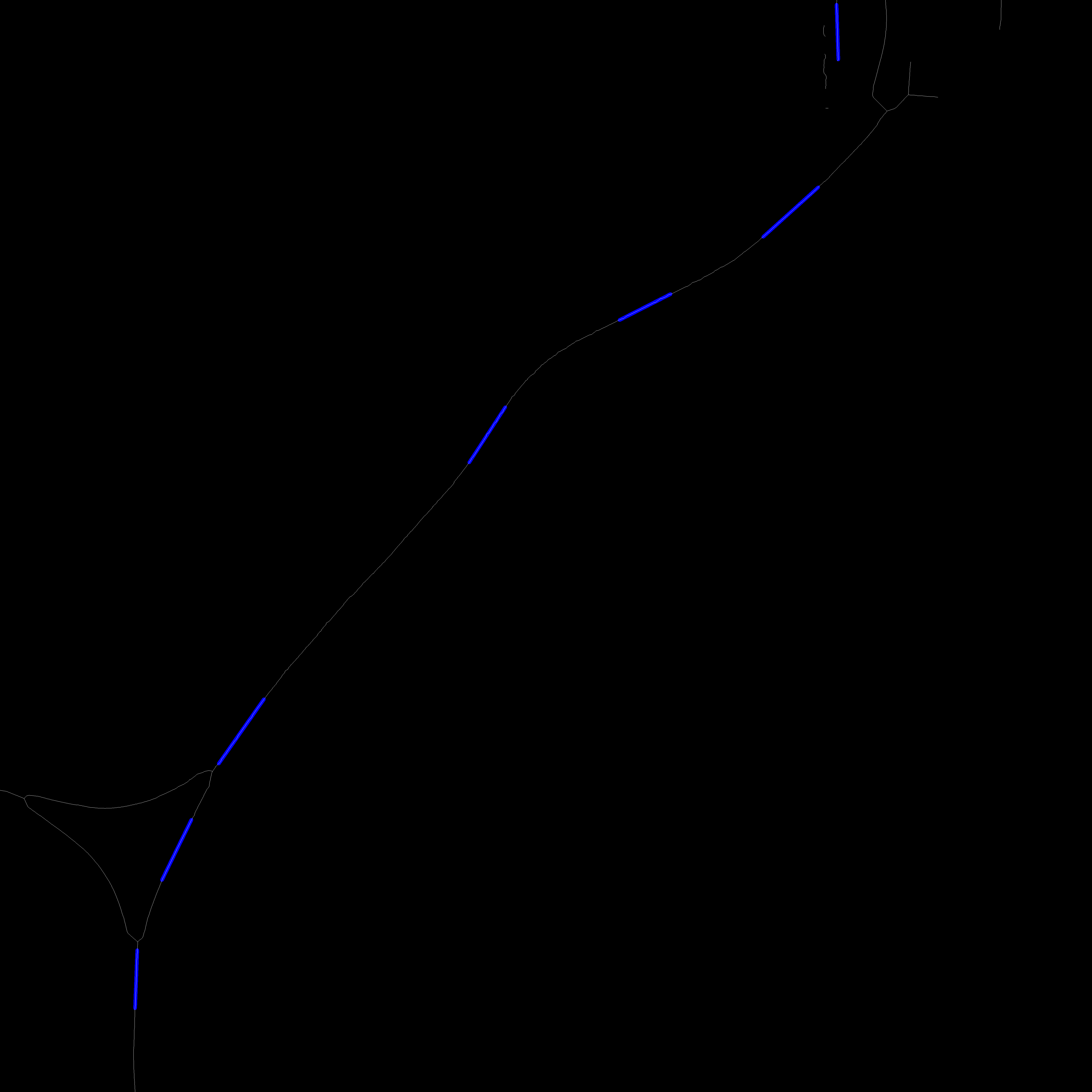}
		\caption{}
	\end{subfigure}
	\begin{subfigure}{0.3\linewidth}
		\includegraphics[width=\linewidth]{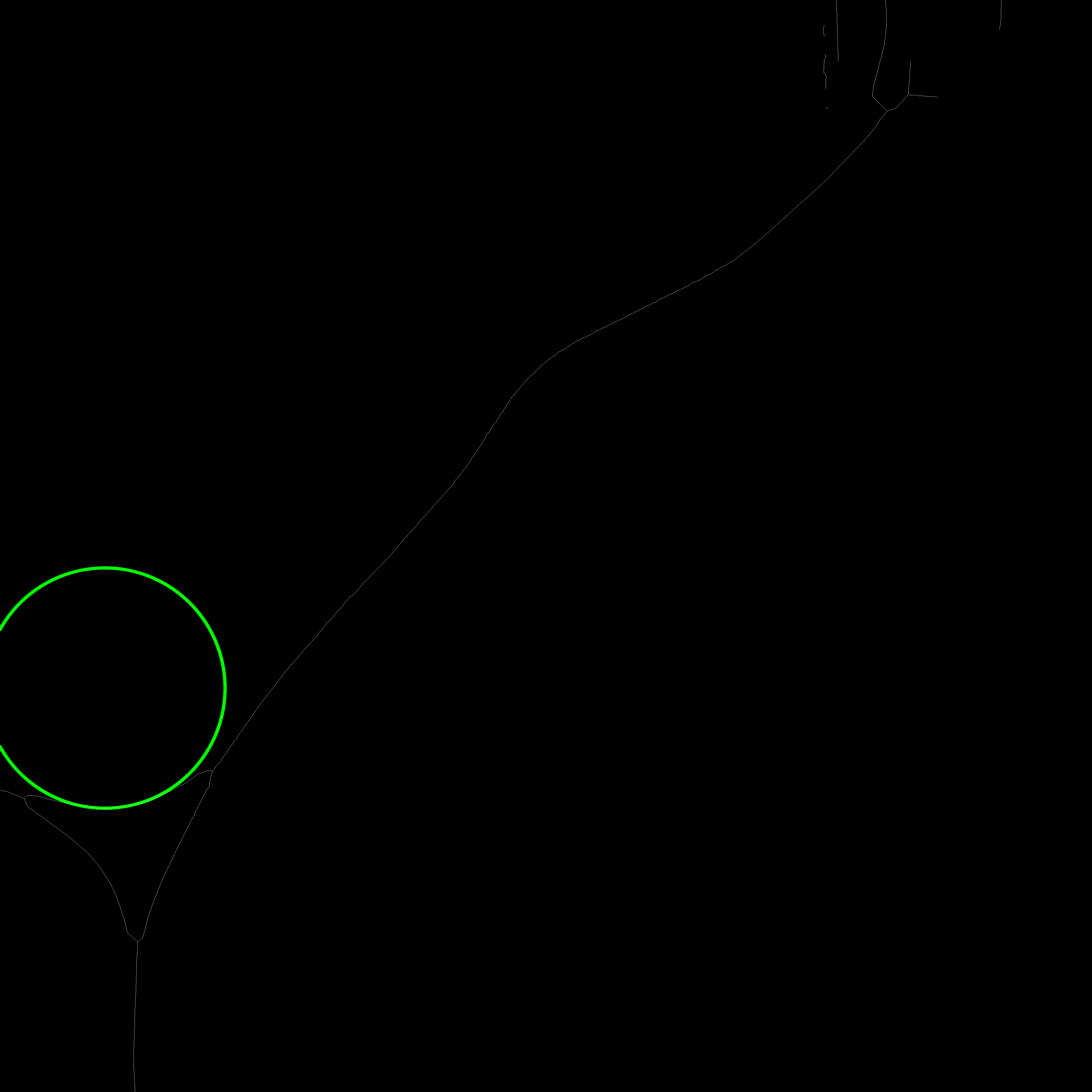}
		\caption{}
	\end{subfigure}
	\begin{subfigure}{0.3\linewidth}
		\includegraphics[width=\linewidth]{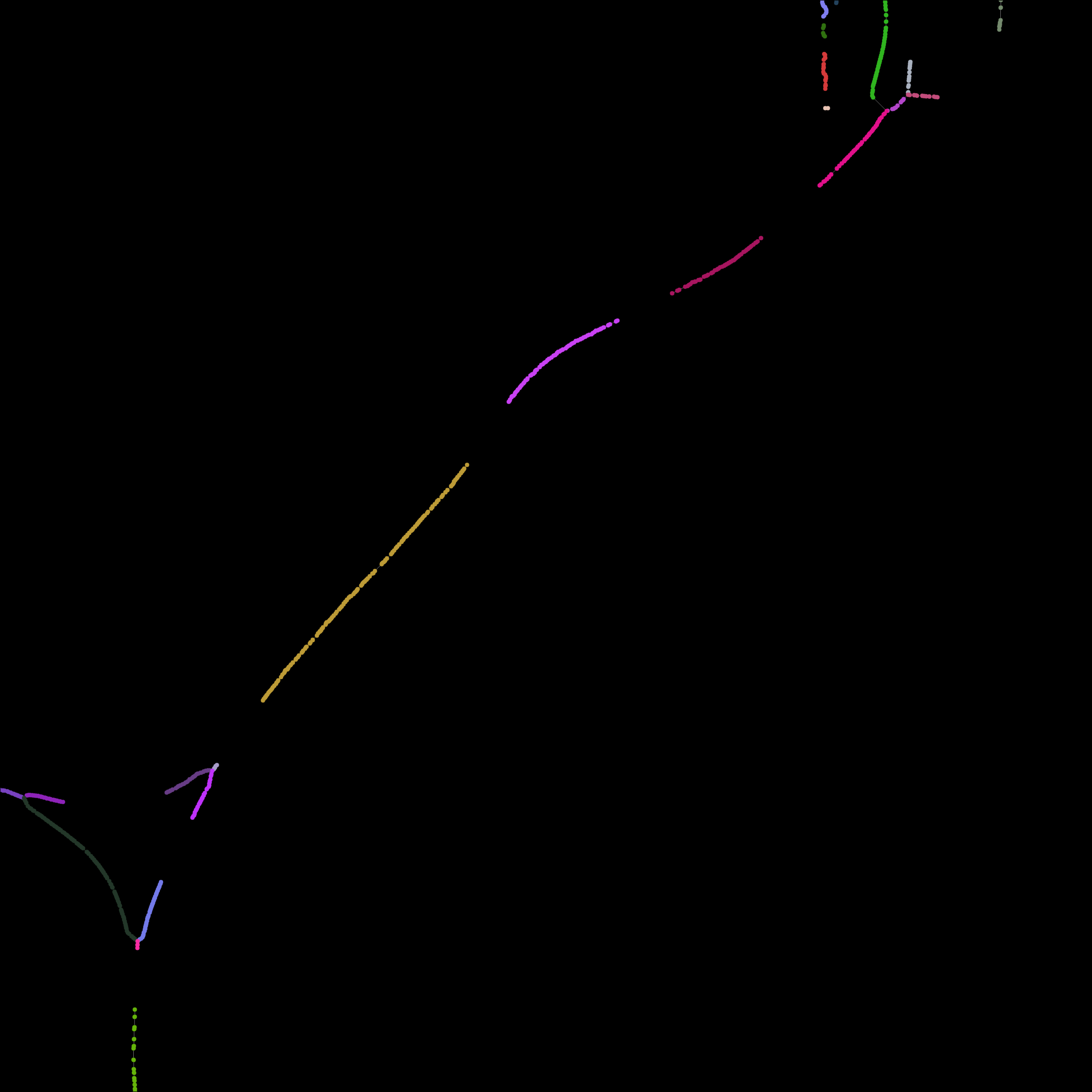} 
		\caption{}
	\end{subfigure}
	\begin{subfigure}{0.45\linewidth}
		\includegraphics[width=\linewidth]{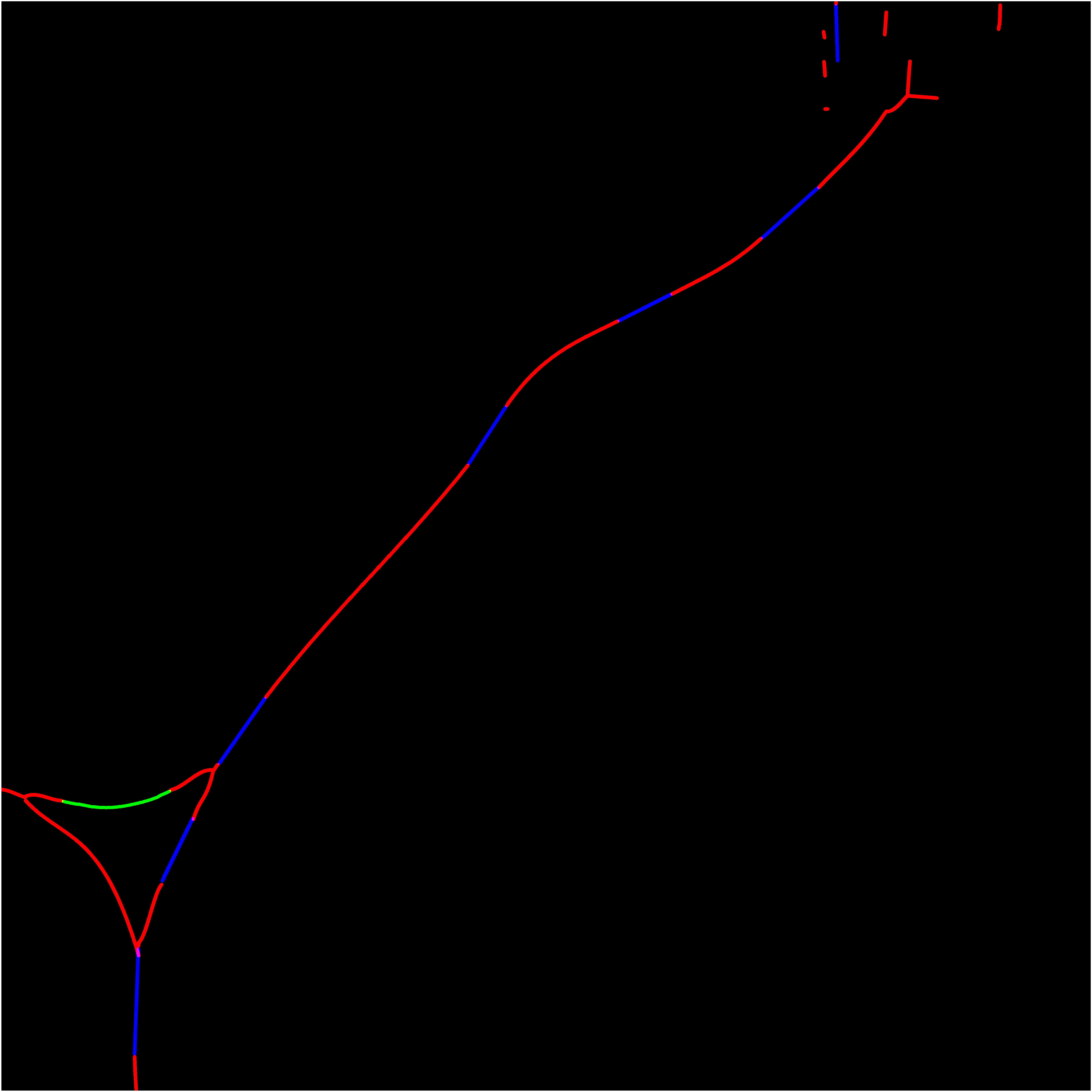} 
		\caption{}
	\end{subfigure}
	\begin{subfigure}{0.45\linewidth}
		\includegraphics[width=\linewidth]{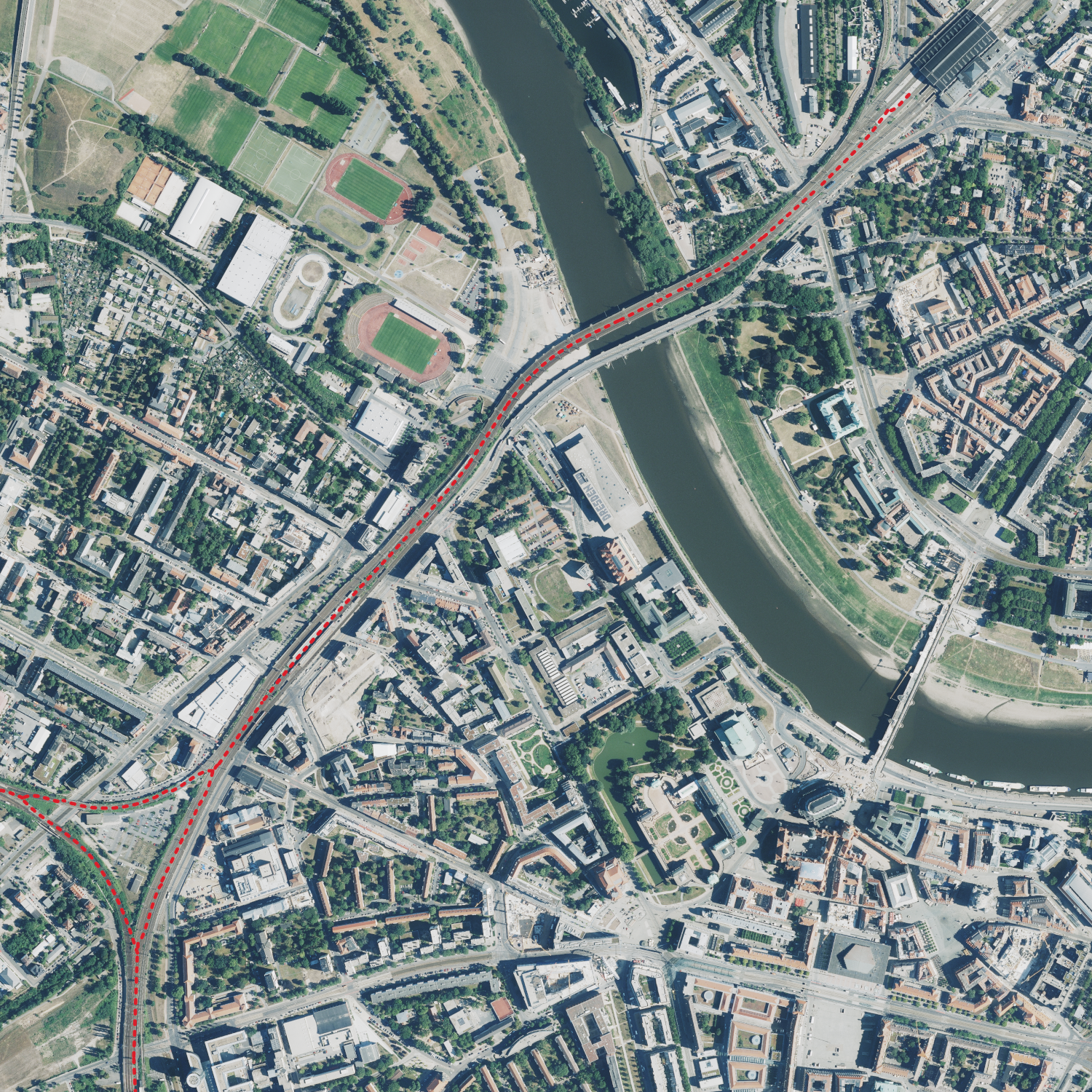}
		\caption{}
	\end{subfigure}
	\caption{Fitting alignment segments to track-bed points; (a) Detected line segments using probabilistic Hough Transform, (b) Best detected Hough circle(s) within the interval [450, 10,000] with 5 metres step, (c) Fitted points to clothoid segments from the decimated remaining curves of the skeletonised image via Visvalingam-Whyatt algorithm (d) Final fitted clothoidal, circular and line segments in red, green and blue respectively, (e) Overlay of the fitted curves and line segments for the recreated railway alignment onto the colourised LiDAR scan viewed in QGIS software.}
	\label{fig:Curve Fitting}
\end{figure}

Seventh, circular segments of the skeleton are similarly estimated with a search interval [450, 10,000] with 5 metres step. The circular segments are extracted from the full circles with a boolean bitwise and operation on the original skeleton then estimated into polar coordinates through the endpoints of every segment to allow for ordered retrieval of each curve's pixel indices from its branch in the minimum spanning tree. Eighth, the remaining skeleton parts are decimated through Visvalingam-Whyatt algorithm and fitted into clothoids with the $G^1$ Hermite interpolation based on the end points and the estimated conformed tangent vector normal to the gradient vector.

Finally, the parameterised line, circular and clothoidal segments are modelled as stringlines in a geodataframe and exported into GeoJSON format to be later used as an input to automatically generate the vertical and horizontal alignments in IFC using IfcOpenShell.

\subsection{BIM of Progressive Textured Meshes}
\begin{figure}[H]
	\centering
	\begin{subfigure}{\textwidth}
		\centering
		\includegraphics[width=\linewidth]{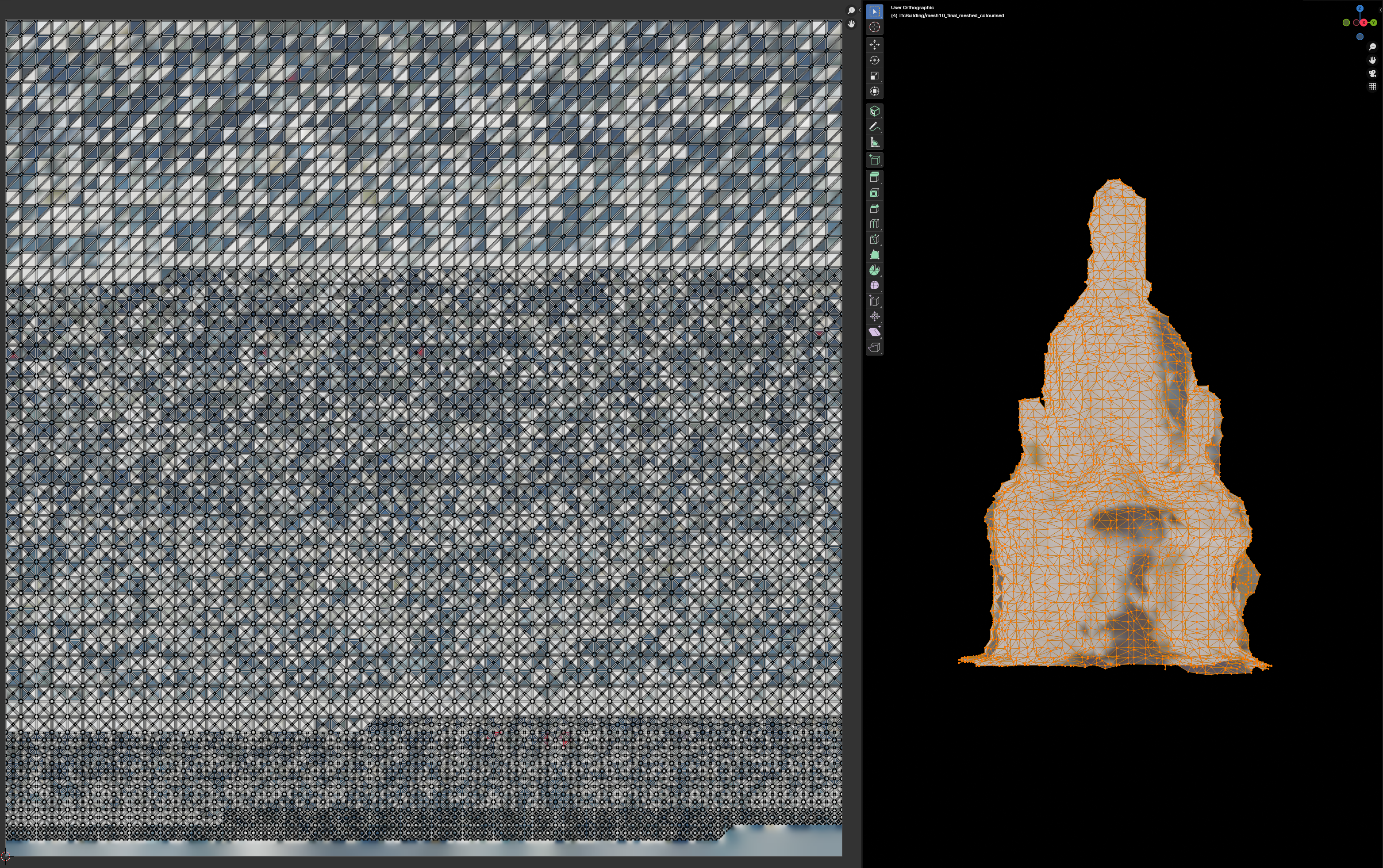}
		\caption{}
	\end{subfigure}
	\begin{subfigure}{\linewidth}
		\includegraphics[width=\linewidth]{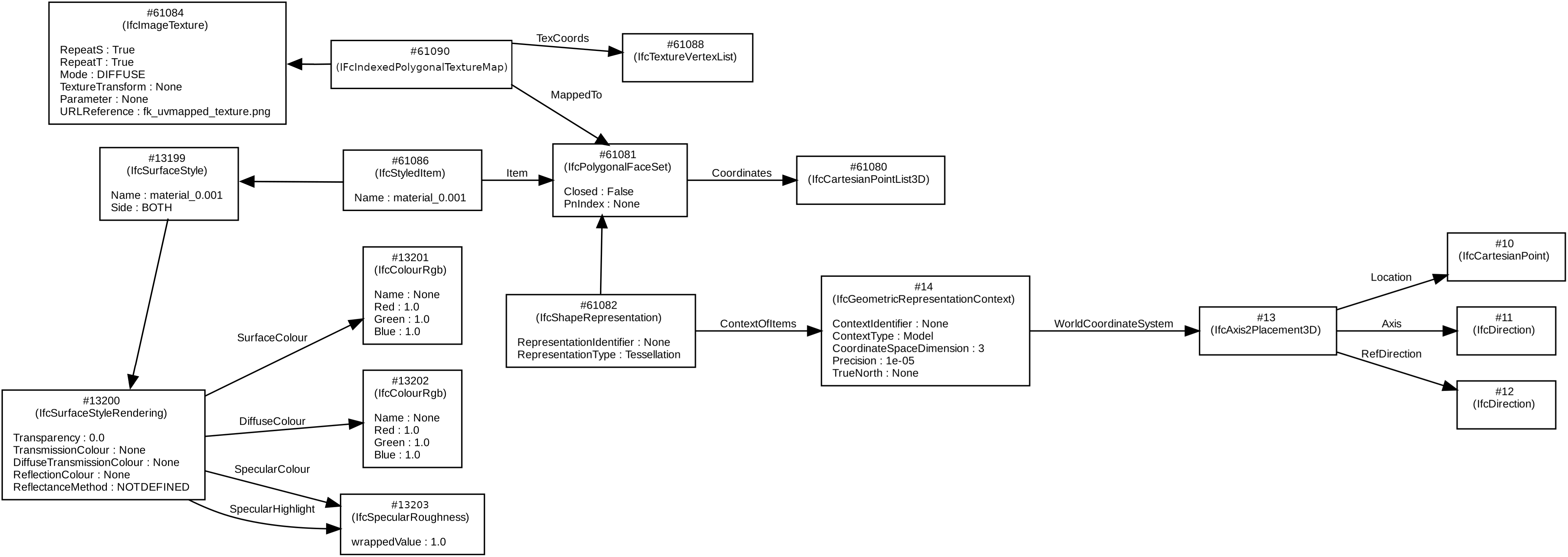}
		\caption{}
	\end{subfigure}
	\caption{A simplified as-is textured IFC model with colours sampled directly from PCD; (a) The textured IFC model on the right and the UV-mapper window on the left displaying the texture atlas as (\textit{IfcImageTexture}) entity automatically mapped to  the shape representation of the triangle meshed surface (\textit{IfcPolygonalFaceSet}) (\textit{IfcSurfaceStyleWithTextures}) through  as If of the \textit{IfcBuilding} instance displayed in BlenderBIM, (b) UML diagram demonstrating the use of \textit{IfcIndexedPolygonalTextureMap} to provide the texture and texture coordinates for \textit{IfcPolygonalFaceSet} from a texture atlas.}
	\label{fig:Frauenkirche_surface_texturing_to_IFC}
\end{figure}

The second use-case is demonstrated on the Church of our Lady (Frauenkirche) found at the lower right corner of the cell in \cref{fig:Frauenkirche_surface_texturing_to_IFC} is used to demonstrate the possibility of integrating 3D coloured texture after meshing and automated UV-mapping of the vertices' colours to a texture atlas into IFC using \href{https://github.com/jpcy/xatlas}{Xatlas}. IfcOpenShell, BlenderBIM and FreeCAD are used to parse the mesh, texture atlas and the UV-map from the Wavefront OBJ format into their relevant IFC entities to generate a simple test project that contains the meshed instance as an \textit{IfcBuilding} and the local path to the texture PNG image linked into the IFC file through a URI. \Cref{fig:Frauenkirche_surface_texturing_to_IFC} shows the IFC model visualised in BlenderBIM alongside the embedded unwrapped UV-Map of the model shown in shown in the UV-editor window on the left. The UML Diagram elaborates the relations defined between the various entities of the textured building.

	\section{Discussion}
\label{}
\subsection{Evaluation of Recreated Alignment}
The difference between the recreated horizontal alignment ($s$) and the provided shape from ATKIS ($t$) using the Root Mean Square Deviation (RMSD) and the coefficient of variation (VC) to the RMSD are calculated using \cref{rmsd,varcoeff}. Over a total length of 3870.444 metres, the RMSD and the VC thereof normalised by the mean value of the measured euclidean distances ($\delta\textsubscript{i}$) are calculated to be $12.506$ metres and $\pm 1.143$ respectively.

\begin{equation}
	\label{rmsd}
	RMSD~(s,t) = \sqrt[2]{\cfrac{\sum_{i=1}^{n} {(s\textsubscript{$ix$}-t\textsubscript{$ix$})^2+(s\textsubscript{$iy$}-t\textsubscript{$iy$})^2}}{n}}
\end{equation}

\begin{equation}
	\label{varcoeff}
	CV~(RMSD) = \cfrac{RMSD}{\widetilde{\delta}\textsubscript{$i$}}
\end{equation}

While the RMSD of fitting the recreated alignment in our presented workflow is higher than already published results in the literature~\cite{Ariyachandra.2021,Cserep.2022}, it still has minimal impact on the precision of consequent output results when deployed for early planning applications. Furthermore, it provides a comprehensible pipeline to define nodes, edges, dependent features, labels and initialisation weights from the CSR Graph of the alignments, which lays the ground for designing a more refined regression solution based on a Graph Neural Network (GNN) framework to be addressed in a future work to derive recreated vertical and horizontal alignments optimally.

\begin{figure}[h!]
	\centering
	\begin{subfigure}{0.45\linewidth}
		\includegraphics[width=\linewidth]{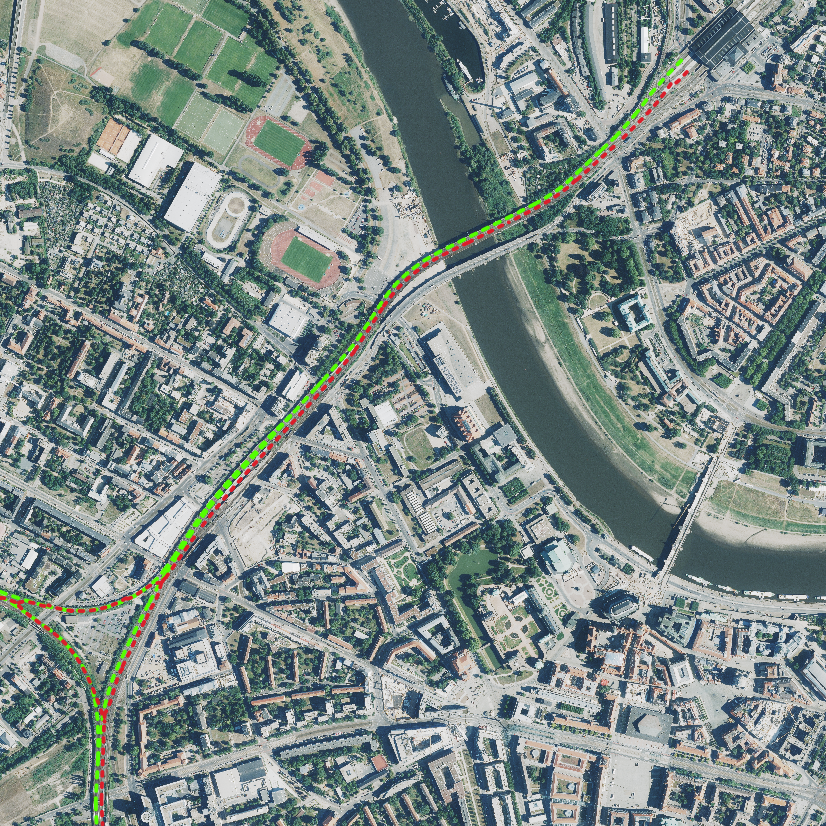}
		\caption{}
	\end{subfigure}
	\begin{subfigure}{0.45\linewidth}
		\includegraphics[width=\linewidth]{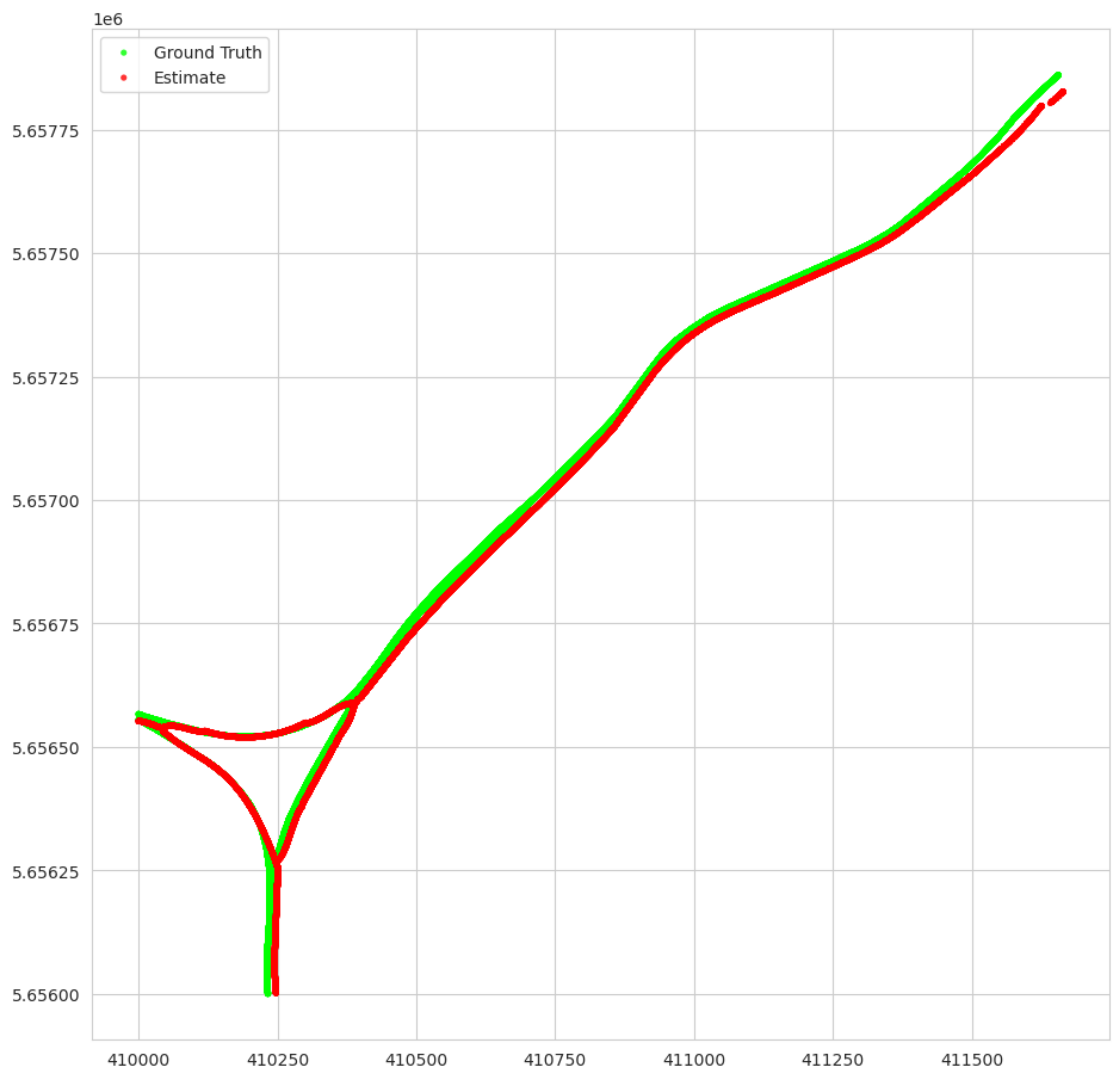}
		\caption{}
	\end{subfigure}
	\caption{Evaluation of the recreated railway alignments; (a) Overlay of the original ATKIS shapes and fitted curves and line segments for the recreated railway alignment in green and red respectively onto the colourised LiDAR scan viewed in QGIS software, (b) A Plot of the original and recreated alignments after recalculating the constructed curves vertices for the same $x\textsubscript{t}$.}
	\label{fig:EVAL_RMSE}
\end{figure}

\begin{figure}[h!]
	\centering
	\begin{subfigure}{0.19\linewidth}
		\includegraphics[width=\linewidth]{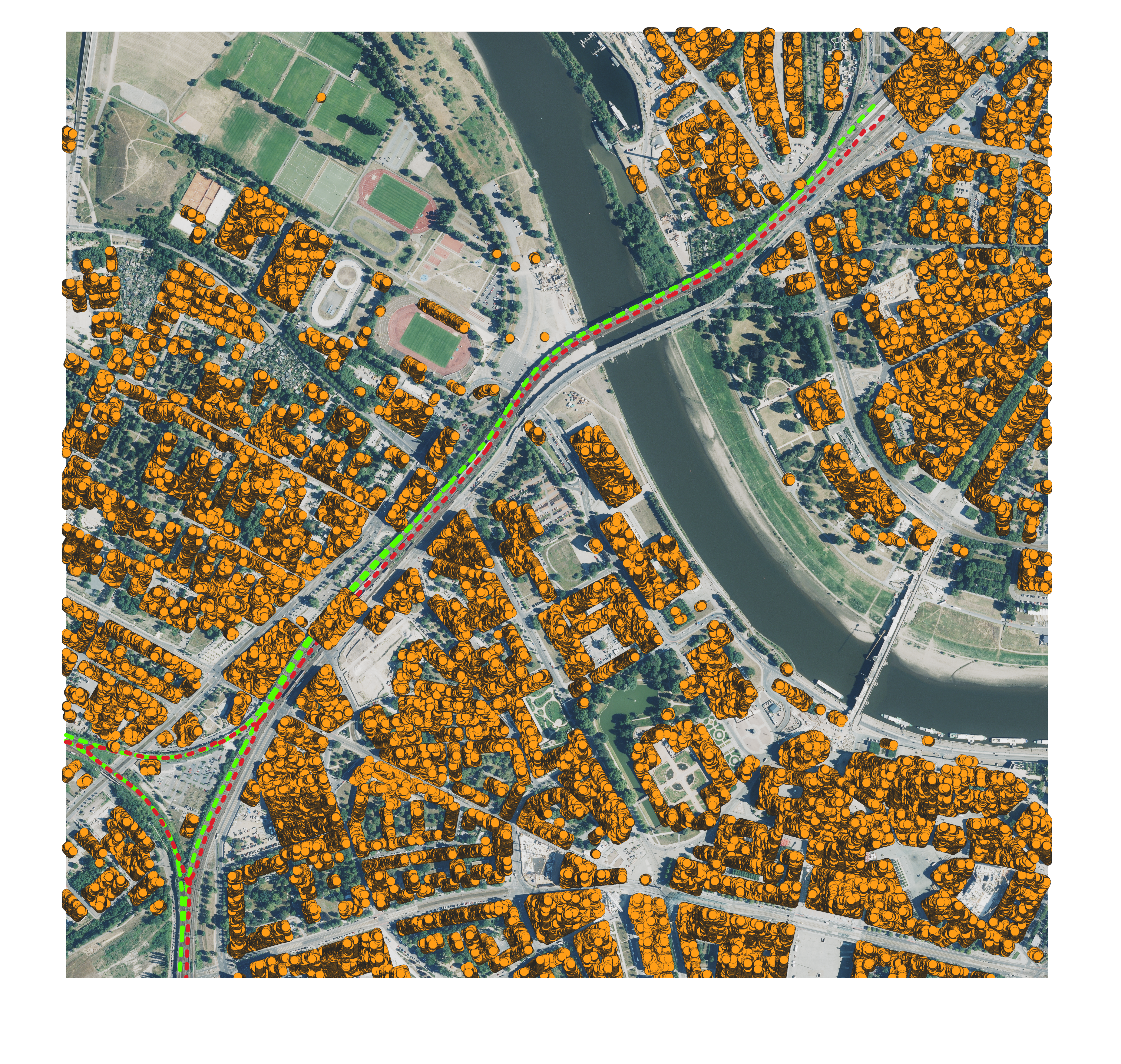}
		\caption{}
	\end{subfigure}
	\begin{subfigure}{0.19\linewidth}
		\includegraphics[width=\linewidth]{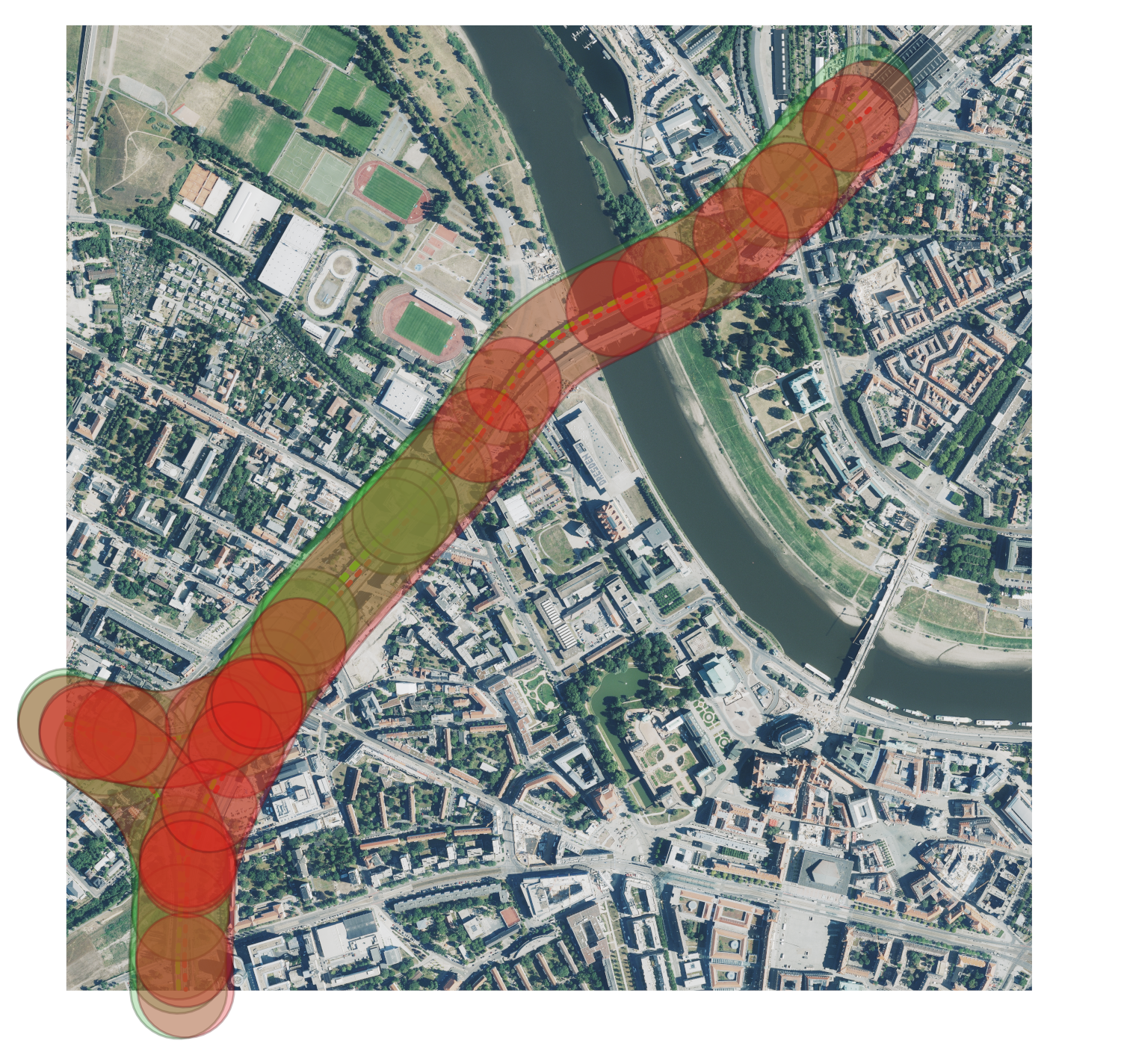} 
		\caption{}
	\end{subfigure}
	\begin{subfigure}{0.19\linewidth}
		\includegraphics[width=\linewidth]{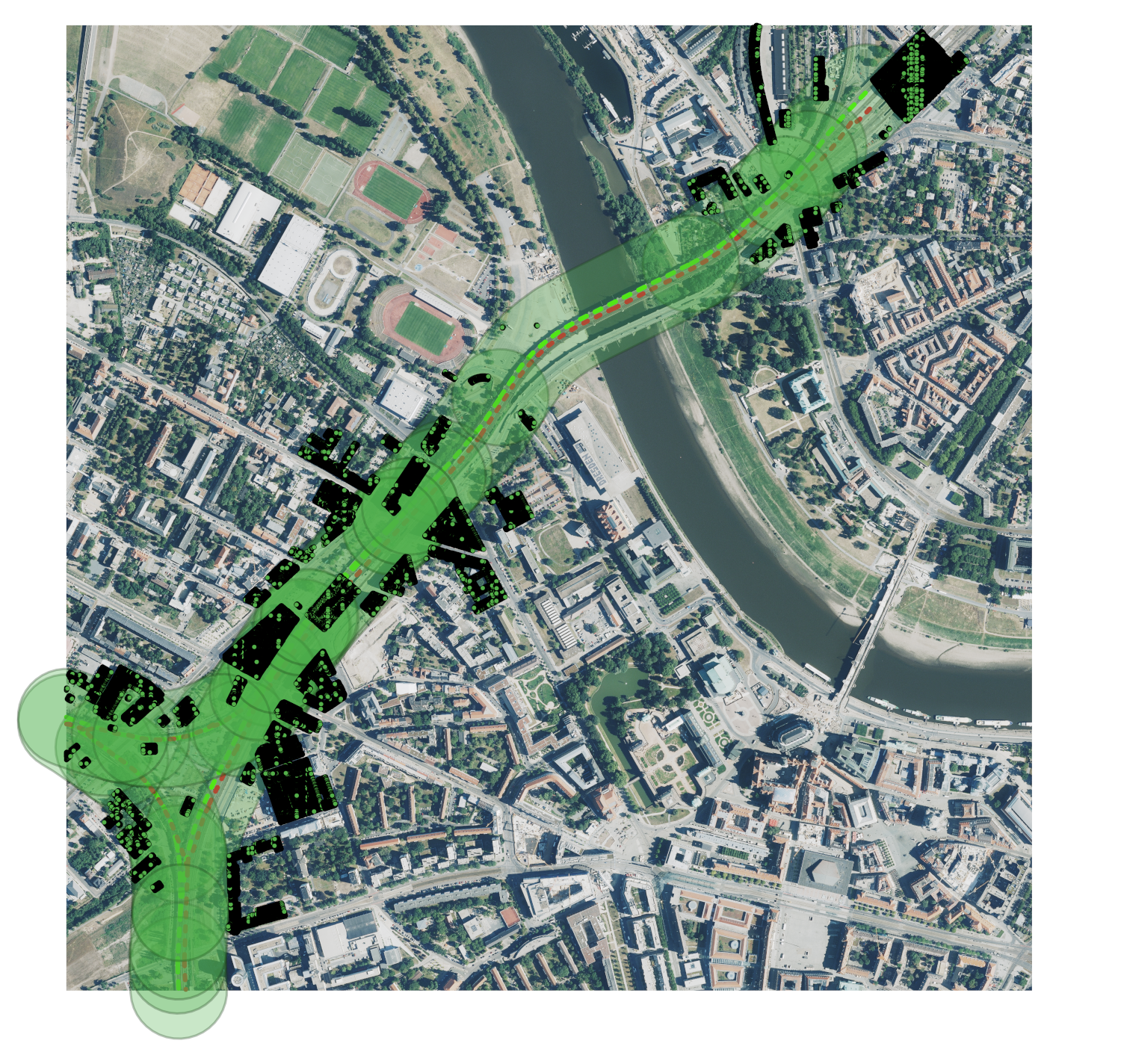} 
		\caption{}
	\end{subfigure}
	\begin{subfigure}{0.19\linewidth}
		\includegraphics[width=\linewidth]{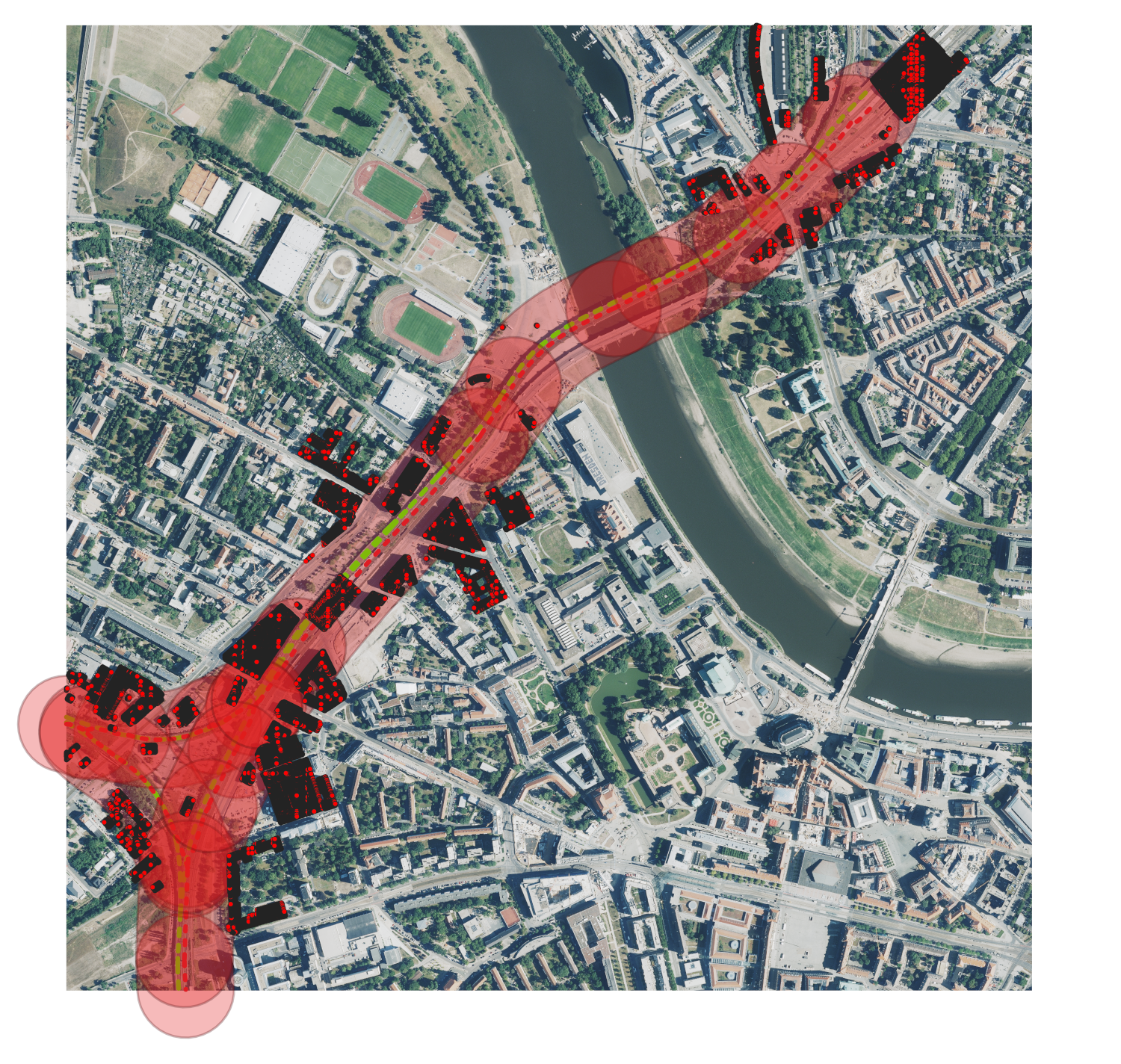}
		\caption{}
	\end{subfigure}
	\begin{subfigure}{0.19\linewidth}
		\includegraphics[width=\linewidth]{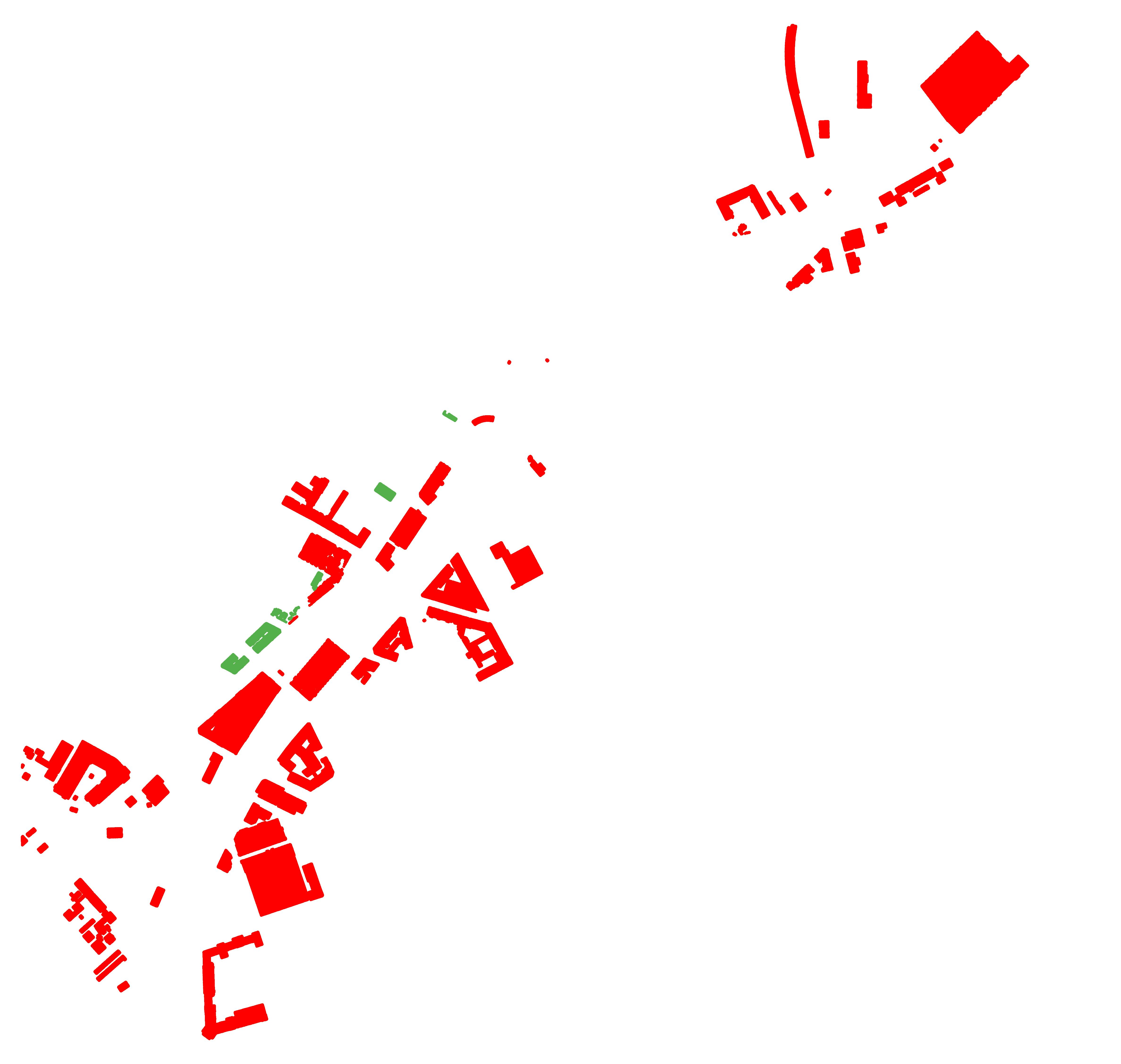}
		\caption{}
	\end{subfigure}
	\caption{Comparison of queried buildings within the vicinity of the recreated railway alignment and the original ATKIS shape using buffer regions of 100 metres; (a) Segmented Buildings PCD displayed onto the colourised LiDAR scan viewed in QGIS, (b) Overlay of the buffer zones of the fitted recreated alignment and the original in red and green respectively, (c) Queried buildings instances that intersect with the buffer zone of the original alignment in green colour, (d) Queried buildings instances that intersect with the buffer zone of the recreated alignment in red colour, (e) Overlay of the queried building instances with a 100 m proximity to the railway alignments from (c) and (d) result in minor difference of around 4.32\% error.}
	\label{fig:Buffering_Comparison}
\end{figure}
\subsection{Limitations}
\subsubsection{Mindful Decision-Making on Technical Feasibility and Applicability}
When striving for very accurate results, as many studies do, the inputs must meet the targeted standard deviation of the results at least. Using less accurate data, by the means of e.g., point cloud density, it must be taken into account that certain details or specific object classes (e.g., overhead wires), can not be accurately reconstructed or are impossible to reconstruct through meshing at all. The selection of object classes to be segmented must therefore be made carefully.
\subsubsection{Accuracy of Colour Sampling from Orthophotos}
Colour value sampling from orthographic photos onto PCD depends on the quality of photos used per se and may vary from previous scans. Orthographic photos from aerial scans do not represent the most accurate colour sampling for vertical elements and planes (e.g., external walls and facades of buildings), as they are technically unfavourable to be captured from above. 

\subsubsection{Inherent Limitations of Flyover LiDAR Scans}
Related to the aforementioned limitation, a well-known fact of reconstruction from scanning is that the results can only be considered accurate for the built environment which is not subterranean. When creating as-is models solely from data from flyovers and/or overground assessments, the bottom edge of the resulting meshes will only represent an assumption and will not take into account subterranean buildings or infrastructure~\cite{Zhang.2023}.

When using PCD and orthographic photos from aerial LiDAR scans, the recorded sets of points differ from others scanned at ground-level, e.g., mobile mapping systems that could be mounted on (rail) vehicles. Rails, overhead wires, signals or other type of built environment, that might be covered by vegetation, adjacent buildings, building components or shadows when using aerial records. The authors aimed at reducing the effect by including GIS data to fill in the gaps, nonetheless, the aforementioned aspect needs to be taken into account and implies, that the trained neural network model might not be suitable for all kinds of PCD, but rather for the ones of aerial origin. The neural network has not been tested on PCD of other nature yet. A possible future improvement could be the combination of aerial and ground-level PCDs to reach a well-balanced dataset and to study the effects that might show in the results.
\subsubsection{Texture Visualisation in IFC Viewer}
The surface reconstruction of segmented PCD into triangle meshes results in a drastic increase in the size of generated shape entities that could leap into the Gigabytes size intervals, if fine meshes and high resolution texture maps are to be used. Most shaders and geometry kernels of the available IFC viewers may not be able to process those on an average daily-use computers due to the limits of disk space, free cache memory and graphics processing capacity.
\subsubsection{Balancing Meshing Complexity with LoD Requirements}
The accuracy resemblance of the reconstructed meshes according to the reality strongly correlates with the accuracy and quality of the input data. Besides that, high geometric and semantic details come with the cost of an increase of size and the amount of information. Therefore, balancing information and LoD needs of as-is models is crucial.
\subsubsection{Advantages and Disadvantages of Texture Embedding in IFC}
The embedding of texture paths as Unique Resource Identifiers (URIs), whether relative or absolute, brings the risk of corrupting the IFC model, if the textures are mislocated or the models are exchanged without their linked attachments. However, entities like \textit{IfcBlobTexture} or \textit{IfcPixelTexture} in the IFC schema provide a workaround by embedding the texture maps directly in the model, albeit at the cost of a larger model size.
\subsection{Potentials}
The demonstrated framework shows the automation of laborious manual as-is modelling, relying solely on openly available point cloud and GIS data. Especially in early or pre-planning phases of infrastructure projects, as-is models hold the potential to support decision-making processes without spending money on costly surveying tasks, leaving more time and budget for continuative thorough analyses. Combined with already established methods of quantitative difference estimation in bi-temporal scans~\cite{James.2017}, relevant queries about changes in context of railway can be collected and further processed to gain relatively accurate and up to date knowledge with minimal additional cost provided that the technical expertise is available. The industry standard IFC makes the results accessible both for further planning based on the created as-is models as well as allowing textured visualisation and documentation in an open format with a suitable data model. Public providers of cadastral and surveying data could, on the same data base they provide now, allocate detailed classified 3D models of the (built) environment to the public. With this, they would enable potential for many possible use cases in the context of infrastructure planning and at the same time relieve the public hand from avoidable expenses.
	\section{Conclusions}
\label{conclusion}
This paper demonstrated a framework for GIS-informed point cloud segmentation resulting in the creation of BIM-ready as-is models of the built environment in railway infrastructure projects. GIS-data was used to create classified masks of coloured point cloud data, including the classes of e.g., vegetation, rail, buildings etc. The segmented point clouds were then used to train a deep learning model, namely 2DPASS, to automatically segment PCD. From the resulting sets of points, class-specific 3D meshes were created, optimised and converted to then openBIM format IFC. The conversion included the proper usage of IFC classes, the enrichment with properties originating from the GIS data and finally, the texturing of the IFC files with aerial images. All data that was used during this process either originated from public surveying data, such as~\cite{ThuringerLandesamtfurBodenmanagementundGeoinformation.2023} or from openly accessible GIS databanks such as OpenStreetMap. 

Future work should consider combining other types of input data, such as 2D plans, with PCD in order to create as-is models. This would fill accuracy gaps and reduce assumptions associated with the use of PCD, such as uncertainties regarding underground structures. The result of such an approach would furthermore reflect more realistically the current work of infrastructure planning by taking into account the best-available data and could contribute to upgrading datasets by contextualising and enriching it. 
	\vspace{6pt} 
	
	\section*{Funding}
	This research received no external funding.
	
	\section*{Data Availability}
	All original GIS data used for the purpose of this study are freely accessible to download and use from the official websites of the Geoportal of the German Free States of \href{https://geoportal.thueringen.de/}{Thuringia} and \href{https://www.geodaten.sachsen.de/}{Saxony} respectively via the Data licence Germany~\textendash~attribution~\textendash~Version 2.0 (\href{http://www.govdata.de/dl-de/by-2-0}{dl-de/by-2-0})
	
	\section*{Acknowledgements}
	{Many thanks to Ms. Judith Krischler and Prof. Christian Koch for their invaluable contribution and expertise in the original EG-ICE 2023 conference paper that greatly assisted this research. Furthermore, we thank Mr. Sergei Rogozin for his Python programming contribution as a student assistant to help retrieve attributes from ATKIS shape files and create buffered labels for annotating some road and rail subclasses in the dataset.}
	
	\section*{Conflicts of Interest}{The author declares no conflict of interest exists that could affect, influence or jeopardise the integrity of research reported in this study.} 
	
	\section*{Abbreviations}{
	The following abbreviations are used in this manuscript:\\
	\noindent 
	\begin{tabular}{@{}ll}
	AEC & Architecture, Engineering and Construction\\
	ATKIS & The Official Topographic Cartographic Information System (German: Amtliches Topographisch-\\
	~ & Kartographisches Informationssystem)\\
	BIM & Building Information Modelling\\
	CCL & Connected Component Labelling\\
	CNN & Convolutional Neural Network\\
	CRS & Coordinate Reference System\\
	DBSCAN & Density-based Spatial Clustering of Applications with Noise\\
	DL & Deep Learning\\
	DNN & Deep Neural Network\\
	EDT & Euclidean Distance Transform\\
	GIS & Geographic Information Systems\\
	IDM & Information Delivery Manual\\
	IFC & Industry Foundation Classes\\
	LiDAR & Light Detection and Ranging\\
	LoD & Level of Development\\
	LoIN & Level of Information Need\\
	MIoU & Mean Intersection over Union\\
	PCD & Point Cloud Data\\
	PNG & Portable Networks Graphics\\
	RANSAC & Random Sample Consensus\\
	SPH & Service Phase\\
	UML & Unified Modelling Language\\
	URI & Unified Resource Identifier\\
	XR & Extended Reality
	\end{tabular}
	}
	
	\bibliographystyle{unsrt}
	\bibliography{bibliography}
\end{document}